\documentclass[journal]{IEEEtran}

\usepackage{cite}
\usepackage{amsmath,amssymb,amsfonts}
\usepackage{algorithmic}
\usepackage{graphicx}
\usepackage{textcomp}
\usepackage{xcolor}
\usepackage{booktabs} 
\usepackage{adjustbox}
\usepackage{gensymb}

\DeclareMathOperator*{\argmax}{arg\,max}

\begin{document}

\title{QFlow: A Learning Approach to High QoE Video Streaming at the Wireless Edge}

\author{\IEEEauthorblockN{Rajarshi Bhattacharyya$^*$, Archana Bura$^*$, Desik Rengarajan$^*$, Mason Rumuly$^*$, Bainan Xia$^*$, Srinivas Shakkottai$^*$, Dileep Kalathil$^*$, Ricky K.P. Mok$^\dagger$, Amogh Dhamdhere$^\dagger$}\\
\IEEEauthorblockA{
	\textit{$^*$Texas A\&M University, College Station, Texas $\quad$ $^\dagger$CAIDA, San Diego, California}} \\
\{rajarshibh, archanabura, desik, masondataminer, xiabainan, sshakkot, dileep.kalathil\}@tamu.edu \\
\{cskpmok, amogh\}@caida.org }

\maketitle

\begin{abstract}
The predominant use of wireless access networks is for media streaming applications.  However, current access networks treat all packets identically, and lack the agility  to determine which clients are most in need of service at a given time.  Software reconfigurability of networking devices has seen wide adoption, and this in turn implies that agile control policies can be now instantiated on access networks.  Exploiting such reconfigurability requires the design of a system that can enable a configuration, measure the impact on the application performance (Quality of Experience), and adaptively select a new configuration. Effectively, this feedback loop is a Markov Decision Process whose parameters are unknown.  The goal of this work is to develop QFlow, a platform that instantiates this feedback loop, and instantiate a variety of control policies over it.   We use the popular application of video streaming over YouTube as our use case.  Our context is priority queueing, with the action space being that of determining which clients should be assigned to each queue at each decision period.  We first develop policies based on model-based and model-free reinforcement learning.  We then design an auction-based system under which clients place bids for priority service, as well as a more structured index-based policy. Through experiments, we show how these learning-based policies on QFlow are able to select the right clients for prioritization in a high-load scenario to outperform the best known solutions with over 25\% improvement in QoE, and a perfect QoE score of 5 over 85\% of the time.

\end{abstract}


\section{Introduction}
\label{intro}

A majority of Internet usage today occurs over wireless access networks, and this trend is only likely to accelerate with the growing penetration of connected televisions, VR headsets, and other smart home appliances.   These access networks are growing ever more dense, and the difference between WiFi and cellular access is becoming less clear as 5G standards that require small, densely located cells, and next generation WiFi standards that utilize per-packet scheduling rather than random access become more popular.  

A major fraction of the packets carried by these wireless access networks are related to media streaming, which have relatively stringent constraints on the required quality of service (QoS) provided by the network for ideal operation.  These QoS metrics typically are measured as link statistics such as $[Throughput,\ RTT,\ Jitter,\ Loss Rate].$  The impact of such QoS on user satisfaction is identified in terms of Quality of Experience (QoE).   QoE is measured as a number in the interval $[1, 5],$ and can be dependent on the application and its evolving state.   For example, the application can be video streaming over the Web, with the state being the number and duration of stalls (re-buffering events) that have been experienced thus far.  Supporting a large number of concurrent streams of this kind, while ensuring high QoE for all clients is a major challenge.

\begin{figure}[htbp]
\begin{center}
\includegraphics[width=3in]{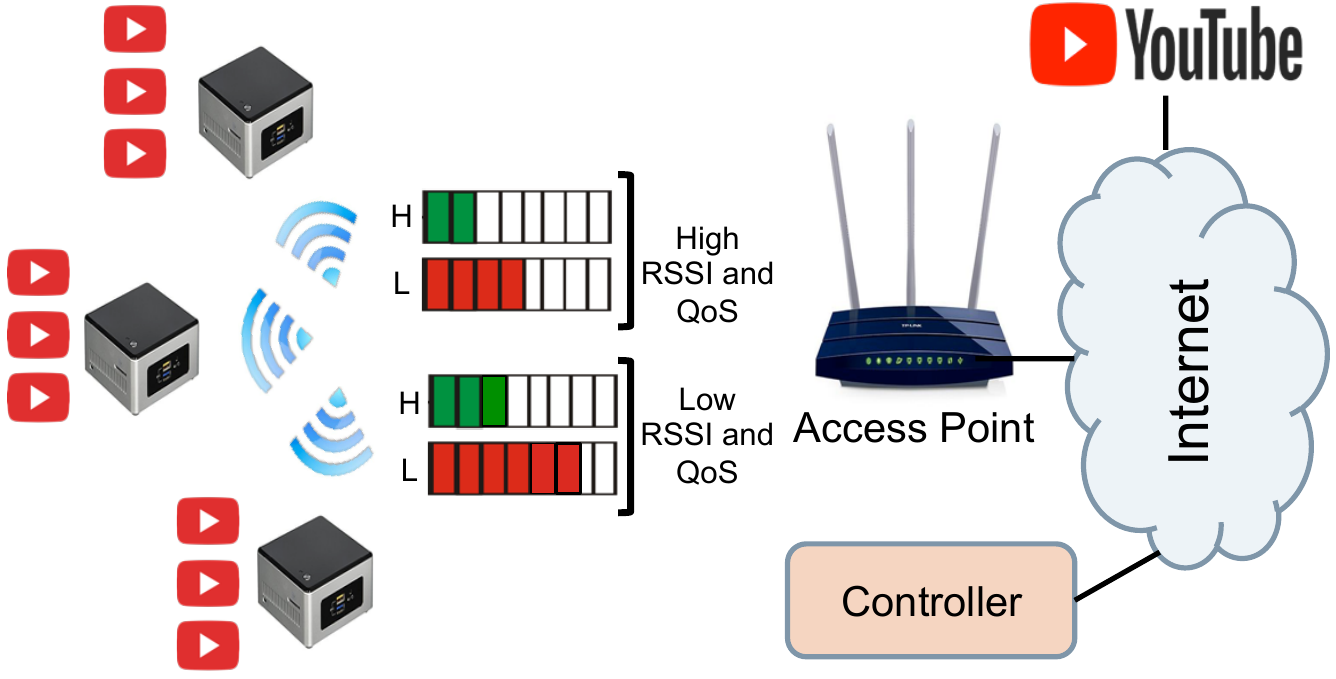}
\caption{Ensuring high QoE video streaming via adaptive prioritization.
}
\label{fig:setup}
\end{center}
\vspace{-0.1in}
\end{figure}
As a concrete example, consider Figure~\ref{fig:setup} that shows 9 simultaneous YouTube clients that are supported over a wireless access network.  This setup is used for our laboratory experiments, and can emulate a range of load and channel conditions by restricting the available QoS values at the access point.   The traditional (vanilla) approach is to maintain a single queue, and to treat all packets identically regardless of the importance of the packets to the QoE of the clients.  So a session that has already buffered up many seconds of video might get equal service as one that is near stalling.  While this approach might be acceptable when the number of streams is limited, the need to support multiple high quality streams motivates the desire to do better.

Given that queuing behavior is fundamental to all elements of the QoS statistics mentioned above, differentiated queuing at the access point immediately suggests itself.  Token-bucket-based shaping can be used to create high-priority and low-priority queues, with the QoS statistics of the former being superior to that of the latter.  Furthermore, we can create multiple ``bins'' of queues as shown in Figure~\ref{fig:setup}, with each bin corresponding to similar client channel conditions (with worse channels implying lower achievable QoS), and allocate them similar time-spectrum resources.   Then a basic question is that of periodically deciding client schedules: \emph{ Given the current QoE and video state at each client, how should the controller assign clients to queues for the next decision period in order to attain system-wide benefits?}

A policy that can attain system-wide benefits requires a feedback control loop of the kind shown in Figure~\ref{fig:loop}.  First, we need to \emph{configure} the system in terms of assigning flows to queues. Second, we need to \emph{measure} the impact of the configuration on QoE \footnote{This is a number in the interval $[1,5]$ that indicates end-user satisfaction, with a QoE of $5$ being the best.} and relevant application state at the end-user.  Third, we need to \emph{learn} what is the relation between realized QoE and the configuration used (using a combination of offline and online learning).  Finally, we need to \emph{adapt} the policy used for configuration as we learn in order to maximize performance goals.   
\begin{figure}[htbp]
\begin{center}
\includegraphics[width=2in]{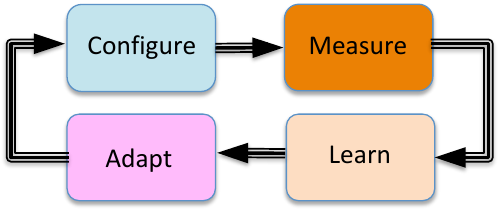}
\caption{Feedback loop for configuration selection.
}
\label{fig:loop}
\end{center}
\vspace{-0.1in}
\end{figure}

Posed in this manner, the application QoE and other measurable application-specific parameters (such as buffered seconds of video) are the observable application state of the system, whose evolution is mediated through the assignment of flows to queues.  The network QoS statistics of each queue are hidden variables that cause transitions to the application state, potentially in a stochastic manner.  The decision of which flows to assign to what queue determines the state transitions that a particular application is exposed to, and must be done in a manner that maximizes QoE.  Thus, the control loop in Figure~\ref{fig:loop} can be interpreted as a Markov Decision Process (MDP) whose transition kernel is unknown, and which could potentially be discovered using reinforcement learning. 

In this work, our goal is to design, implement, and evaluate QFlow, a platform for reinforcement learning that instantiates the feedback control loop described above on a WiFi access point that faces a high demand.  Performance over high capacity wired backhaul links is near-deterministic, and resources constraints apply tot he last hop wireless link.  We choose video streaming as the application of interest using the case study of YouTube, since video has stringent network requirements and occupies a majority of Internet packets today \cite{ericsson-mobility-report}.




\subsection*{Main Results}



\textbf{Queue Configuration:}  We enable reliable delivery of configuration commands to hardware that can support re-configuration.
We extend the OpenFlow protocol (currently limited to the network layer) in a generic manner that enables us to use it reconfigure queueing mechanisms.  We select commercially available WiFi routers with Gigabit ethernet backhaul as the wireless edge hardware.  Reconfigurable queueing is attained by leveraging differentiated queueing mechanisms available in the Traffic Controller (tc) package by installing OpenWRT (a stripped-down Linux version).  Here, we can choose between queueing disciplines and set filters to assign flows to queues.  Details are presented in Section~\ref{section:softstack}.

\textbf{Measurement of Application State and QoE:}  We enable continuous monitoring of client-specific application state consisting of buffered seconds of video and stall duration (when the video re-buffers).  These monitors at the WiFi router and the mobile station, are compatible with our OpenFlow extensions, and use the protocol to periodically send statistics to the OpenFlow controller for processing.    We continuously predict the QoE of the ongoing application (video streaming) flows as a function of the application state using existing maps of the relationship between video events (such as stalls) and QoE.  Details are presented in Sections~\ref{section:system},~\ref{section:softstack}.

\textbf{Model-Free Reinforcement Learning:} We  develop a model-free reinforcement learning (RL) method that enables adaptation to the current QoE and application state over all users to maximize the discounted sum of QoEs.  We design a simulator that approximates the evolution of the underlying system, and its impact on application state and QoE.  We use the simulator to train a Q-Learning algorithm, with non-linear function approximation using a neural network.   This so-called Deep Q Network (DQN) is able to account for state space explosion across the users and provides a Q-function approximation for all states.  Details are presented in Section~\ref{section:ModelFreeRL}.

\textbf{Model-Based Reinforcement Learning:} We next develop a model-based RL approach based on the observation that the state evolution of an individual client is independent of  others given the action (queue assignment).  We first use measurements conducted over the system using a range of control policies to empirically determine the transition probabilities on a per-client basis, and then use the independence observation to construct the system transition kernel (this applies to the vector of all client states taken together).  While doing so, we reduce the system state space by discretization and aggregation to a subset of frequently observed system states.  Finally, we solve the MDP numerically to obtain the model-based policy.

\textbf{Auction:}
%
The approaches above require that the state of each client be supplied by the clients themselves, which implies that strategic clients could obtain more than their fair share of resources through appropriate state declarations to the AP.  We hence consider 
an incentive compatible (truth-telling) auction for ensuring high QoE video streaming. Here, bids for the auction are placed via a smart middleware algorithm \emph{(not by a human end-user, who may be unaware of the existence of the system)}, and can be interpreted as the number of cents that the bidding algorithm is willing to pay for high priority service for the next 10 seconds\footnote{We calculate that the eventual dollar price paid will be of the order to a few tens of dollars per month, consistent with cellular data access billing schemes of today.}.  The agents are provided the model (transition kernel) as in model-based RL, as well as the empirical bid-distribution of all the agents, and use these to obtain the best response bid, consistent with our earlier work on a mean field game approach to scheduling~\cite{ManRam14,manjrekar2019mean}.  The per-agent bid computation under this regime is straightforward, and details are presented in section~\ref{section:auction}.

\textbf{Index Policy:}  The results from the auction approach suggest that a indexing of state in the manner of the Whittle index~\cite{whittle1988restless} is possible, under which each client state is associated with a  real-number index.  The optimal policy simply picks the clients with largest indices to promote to the high priority queues.  We empirically validate this hypothesis by using the value function of a given state derived from the auction as its index,  and find that such an index policy performs as well or better than all others, lending credence to the indexing claim.  Details are presented in Section~\ref{section:index-policy}.

\textbf{Experimental results:} The experimental configuration consists of a single queue in the base (vanilla) case, and two reconfigurable queues in the adaptive case.   We conducted experiments in both a static scenario of 6 clients, as well as a dynamic one in which anywhere between 4 and 6 clients are in the system at a given time.  Apart from auction-based, model-free and model-based RL, we also implemented round-robin assignment, greedy maximization of expected QoE, and greedy selection of the clients with lowest video buffers (this policy has been shown to ensure low probability of stalling \cite{singh2015optimizing}).  Our results on adaptive flow assignment (Section~\ref{section: experiments}) reveal that the vanilla approach of treating all flows identically has significantly worse average QoE than adaptive approaches.

Interestingly, the model-based, model-free and auction-based approaches ensure that any given client experiences a perfect QoE of 5 over 85\% of the time, whereas the best that any other policy is able to achieve is only about 60\%, while vanilla manages even less at about 50\%.  This impressive performance improvement of about 25-30\% indicates that by selecting flows in need of QoE improvement (due to high likelihood of stalls in the near future), RL-based adaptive flow assignment improves QoE for the majority of clients.

An earlier version of this work appeared in \cite{bhattacharyya2019qflow}, in which we presented the basic model-based and model-free RL approaches operating over the QFlow platform.  The differences between our earlier work and this paper are as follows: (i) we develop an auction platform for clients to compute and bid their perceived valuations (under the transition kernel generated by model-based RL), and show empirically that it attains similar (slightly better) performance than the two RL approaches, (ii) we explore the idea that policies can have structure, and show empirically that a simple index-type of policy might be optimal for the class of flow prioritization problems that we consider, and (iii) we show empirically that the indices (state ordering by value) developed for a larger number of clients follow a similar order to those for a smaller number of clients, meaning that a single set of indices work well without retraining, even under a dynamically changing number of clients with time varying channel conditions.


\section{Related Work}
\label{section:related}

\emph{Optimal Queueing:} There has been significant work on QoS as a function of the scheduling policy, e.g., a sequence of work starting with \cite{TasEph_92}, and follow on work in the wireless context that resulted in algorithms such as backpressure-based scheduling and routing in wireless networks \cite{ErySriPer_05} and more recently \cite {HouBor09} that ensures that strict delay guarantees are met.   Most of these works aim at maximizing throughput or loss rate, but they do not consider all the elements of QoS together. Also, they do not map received QoS to application QoE.

\emph{SDN-based Video Streaming:}
A number of systems have been proposed to improve the performance and QoE of video streaming with SDN. One direction is to assign video streaming flows to different network links according to various path selection schemes \cite{jarschel13sdnyoutube} or the location of bottlenecks detected in the WAN \cite{nam14qoesdn}. In the home network environment, the problem shifts from managing the paths of video traffic to sharing the same network (link) with multiple devices or flows. VQOA \cite{ramakrishnan15sdnqoe} and QFF \cite{Georgopoulos13openflowqoe} employ SDN to monitor the traffic and change the bandwidth assignment of each video flow to achieve better streaming performance. However, without an accurate map of action to QoE, the controller can only react to QoE degradation passively.

\emph{Reinforcement Learning:}  An RL approach is natural for the control of systems with measurable feedback under each configuration.  {The idea of using RL in the context of video streaming rate selection has been  explored in \cite{mao2017neural,huang2018qarc,zhang2019drl360,xiao2019deepvr}. } Different control theoretic methods, such as model predictive control \cite{yin2015control} and PID control \cite{qin2019control} have also been used for adaptive video streaming.   This body of work can be seen as the complement of our own.  Whereas  we are interested in allocating network resources (at the wireless edge) to suit concurrent video streams, their goal is to choose the streaming rate to suit the realized network characteristics. 
Finally, deep reinforcement learning algorithms have been used to solve a number of problems in communication network applications, although not in our problem space; see  \cite{luong2019applications} for a survey. 

\emph{Auctions and Scheduling:} There has also been work on using price or auction-based resource allocation in the wireless context. On the analytical side, \cite{Auction06} considered the problem of auction-based wireless resource allocation.  Here, users participate in a second price auction and bid for a channel.  It was shown that with finite number of users, a Nash Equilibrium exists and the solution is Pareto optimal.  In \cite{ManRam14,manjrekar2019mean}, an auction framework is presented in which queues (representing apps on mobile devices) repeatedly bid for service in a second-price auction that determines which set of queues will be selected for service.  They show that under a large system scaling (called the mean field game regime), the result of the auction would be the same as that of the longest-queue-first algorithm, and hence ensuring fair service for all.  Our design of auction-based scheduling algorithms are motivated by these ideas.  In the context of experiments, a recent trial of a price-based system is described  in \cite{HaSen12}.  Here, day-ahead prices are announced in advance to users, who can choose to use their cellular data connection based the current price.   Thus, the decision makers are the human end-users that essentially have an on/off control.   Furthermore, the prices are not dynamic and have to be determined offline based on historical usage.

\emph{OpenFlow Extensions:} There has been significant research into the development of OpenFlow extension to cross-layer wireless configuration selection.  In this context, CrossFlow \cite{CrossFlow1,CrossFlow2} uses the SDN framework for configuring software defined radios.  Similarly \AE therFlow~\cite{Muxi}, extends OpenFlow for enabling remote configuration of WiFI access points.  Finally, recent systems such as  AeroFlux \cite{aeroflux} and OpenSDWN \cite{schulz2015opensdwn}  enable packet prioritization for flows that are identified by packet inspection as belonging to high priority applications, such as video streaming.  However, these are all offline static policies in that they do not relate the prioritization policy with the state of the application.

\section{System Model and Architecture}
\label{section:system}
We consider a system in which clients are connected to an wireless Access Point (AP) in a high demand situation. We choose video streaming as the application
of interest using the case study of YouTube, since video has stringent network requirements and occupies a majority of Internet packets today \cite{ericsson-mobility-report}. Our goal is to maximize the overall QoE of all the clients in this resource constrained situation.  

The AP has a high priority and low priority queue.  Here, we mean that clients assigned to the high priority queue typically  experience a better QoS (higher bandwidth, lower latency etc.) when compared to the clients assigned to the low priority queue.  The controller assigns clients to each of these queues at every decision period (DP; 10 seconds in our implementation).  Determining the optimal strategy is complex, since the controller does not have prior knowledge of the system model.  Hence, the controller must \textit{learn} the system model and/or control law.  

\subsection{Markov Decision Process}
We consider a discrete time system where time is indexed by $t \in \{0,1,...\}$.  At each DP ($t=0,1,2..$) the controller makes an assignment of clients to queues, and observes the system.  Based on its observation and previous assignment, the controller makes an assignment in the next DP, eventually learning the system model empirically.   This class of problem falls within the Reinforcement Learning (RL) paradigm, and thus can be abstracted to a general RL framework consisting of an \textit{Environment} that produces \textit{states} and \textit{rewards} and an \textit{Agent} that takes \textit{actions}.

\textbf{Environment:} The environment is composed of clients and the AP.  Let $\mathcal{C}$ denote the set of clients.  

\textbf{State:}  Each client keeps track of its state which consists of its current buffer (the number of seconds of video that it has buffered up), the number of stalls it has experienced (i.e., the number of times that it has experienced a break in playout and consequent re-buffering), and its current QoE ( a number in $[1,5]$ that represents user satisfaction, with 5 being the best).  The state of the system is the union of the states of all clients.  Let $s^c_t$ denote the state of client $c$ at time $t$ and $s_t$ denote the state of the system,
\begin{align*}
	&s_t^c =\textit{[Current Buffer State, Stall  Information, Current QoE] } \\ &\hspace{220pt}\forall c \in \mathcal{C} \\
	&s_t=\left[\cup_{\forall c \in \mathcal{C}} s_t^c\right]
\end{align*}

\textbf{Scheduler Action:} The scheduler is the agent that takes queue assignment actions in every decision period in order to maximize its \textit{expected discounted reward}.  Let $a_t^c \in \{0,1\}$ denote the action taken on client $c$ at time $t,$ where $1$ and $0$ indicate assignment to the high and low priority queue, respectively.  Let the set of overall actions be denoted $\mathcal{A}.$  Each such overall action is of form $a_t = [a_t^1, a_t^2, \cdots a_t^{|\mathcal{C}|}].$  The scheduler may assign only $N$ clients to the high priority queue, i.e., $\sum_{c \in \mathcal{C}} a_c^t = N.$  


\textbf{Reward:} The per-client reward $R(s_{t}^{c},a_{t}^{c})$ resulting from taking action $a_t$ at state $s_t$ is the QoE of client $c$ in state $s_{t}^c$.  The overall reward $R(s_t,a_t)$ is the average QoE of all clients in state $s_{t+1},$
$$
R(s_t,a_t) = \frac{1}{|\mathcal{C}|}\sum_{c \in \mathcal{C}} R(s^{c}_{t}, a^{c}_{t})
$$



\textbf{Transition Kernel:} Let ${P}(s_{t+1}|s_t,a_t)$ denote the system transition kernel. 

\textbf{Policy:} The goal of the agent is to maximize the overall QoE of the system.  This goal can be formulated as maximizing the expected discounted reward over an infinite horizon.  Let $\pi(a_t|s_t)$ denote the probability of taking action $a_t$ given the current state (called the policy) and $\gamma$ denote the discount factor. Then the goal is to find $\pi^*$, the policy that maximizes the expected discounted reward,  
\begin{equation} 
\pi^*=\text{argmax}_{\pi} \mathbb{E}\left[\sum_{t=0}^\infty \gamma^t R(s_t,a_t)|s_0=s,a_t \sim \pi(\cdot|s_t)\right].
\end{equation}


\subsection{Auction}
We consider a market wherein clients bid for high priority service periodically. In each discrete time instant, a fixed number of clients $N$ are assigned to the high priority queue. Clients participate in an $(N+1)^{th}$ auction to compete for admission to the high priority queue. The $N$ winners who obtain high priority service will pay a price that is equal to the $(N+1)^{th}$ highest bid, and the rest of the clients will be assigned to the low priority queue. We model the system in a Mean Field approach as described below,

{\bf{Bid}:} The bid submitted by the client in each auction is denoted by $b\in\mathcal{B}$, where $\mathcal{B}$ is a set containing discrete bid values.  The bids can be seen as the price each client $c$ is willing to pay to obtain high priority service.  Note that the human end user plays no role in selecting these bids.

{\bf{Bid Distribution}:} The clients must place their bid based on the beliefs of their competitors. We denote the assumed bid distribution in the market as $\rho$. 

{\bf{Auction Outcome}:} The probability that a client $c$ wins at the auction depends on the event that its bid is greater than the $N+1^{th}$ bid.  Under the mean field approximation, the client assumes that competing bids are all drawn in an IID manner from $\rho.$   We denote the probability of winning under this assumption when a client places a bid $b$ by $p_{win}(b),$ where we have dropped the dependence on $\rho$ for ease of notation.  Thus, $p_{win}(b),$ is the probability that $b$ lies within the top $N$ values of $|\mathcal{C}|-1$ independent draws from $\rho.$  Accuracy of the mean field approximation in the regime where there is a large pool of clients from which a small subset is drawn at each auction is available in \cite{ManRam14,manjrekar2019mean}   

{\bf{Payment}:} The amount paid after each auction is denoted by $pay$. Note that $pay$ is a random variable that corresponds to the auction mechanism. In particular, the payment distribution in our system upon winning is exactly the distribution of the $(N+1)^{th}$ highest bid.

{\bf{Scheduler Action:}}  As in the previous case, the scheduler decides on which clients to assign to each queue.  However, here the actions are taken based on the outcome of the auction.  The actions $a_t^{c}=1$ and $a_t^{c}=0$ correspond, respectively, to winning and losing at the auction by the client $c.$  

{\bf{Client Reward}:} 
As before, reward $R(s_{t}^{c},a_{t}^{c})$ resulting from  action $a_t^c$ at state $s_t^{c}$ is the  QoE of client $c$ in state $s_{t}^c$.  Note, however, that each agent is only concerned with its own reward, unlike the average case discussed earlier.

\textbf{Client Transition Kernel} Let ${P}(s_{t+1}^{c}|s_t^{c},a_t^{c})$ denote the client transition kernel.  Thus the probability of transitioning to state $s^{c}_{t+1}$ is jointly defined by the probability of winning the auction when bidding $b$, $p_{win}(b)$ and  ${P}(s^c_{t+1}|s^c_t, a^{c}_t)$. 

{\bf{Policy}:} 
%
%
The agent (client) must place a bid at each time, accounting for its progression of state.  Following the same methodology as \cite{ManRam14, manjrekar2019mean}, we formulate the optimal policy (bid decision) problem of the corresponding Markov Decision Process as follows: 
\begin{align}
&b^{*}(s^c_t)= 
\text{argmax}_{b\in\mathcal{B}} \bigg\{ p_{win}(b)\Big[R(s^c_{t+1}, a^{c}_t=1) - pay + \nonumber\\ 
&\sum_{s^c_{t+1}} {P}(s^c_{t+1}|s^c_t,a^{c}_t= 1)\gamma v(s^c_{t+1})\Big] +\nonumber\\
&  (1- p_{win}(b))\Big[R(s^c_{t+1},a_t^{c}=0) + \nonumber\\
&\sum_{s^c_{t+1}}{P}(s^c_{t+1}|s^c_t,a_t^{c}= 0)\gamma v(s^c_{t+1})\Big] \bigg\},
\label{eq:val}
\end{align}
where $v(.)$ is the optimal client value function.

\subsection{Measuring QoE for Video Streaming}

Considerable progress has been made in identifying the relation between video events such as stalling, and subjective user perception (QoE) \cite{7025402,8013810,8247250} via laboratory studies.
However, these studies are insufficient in our context, since they do not consider the network conditions (QoS statistics) that gave rise to the video events.  
Nevertheless, we can leverage these studies by using them as models of human perception of objectively measurable video events.  
We considered three models in this context, namely Delivery Quality Score (DQS) \cite{7025402}, generalized DQS \cite{8013810}, and Time-Varying QoE (TV-QoE) \cite{8247250}.  All of the three models are based on the same features (stall event information) if there is no rate adaptation.  Since our goal is to support high resolution video without degradation, we fix the resolution so as to prevent video rate adaptation.  Under this scenario, all three models are similar, and we choose DQS as our candidate.  Note that DQS has been validated using 183 videos and 53 human subjects \cite{7025402}, and we do not repeat these experiments. 


\begin{figure*}[htbp]
\centering
\begin{minipage}{.26\textwidth}
\centering
\includegraphics[width=0.9\columnwidth]{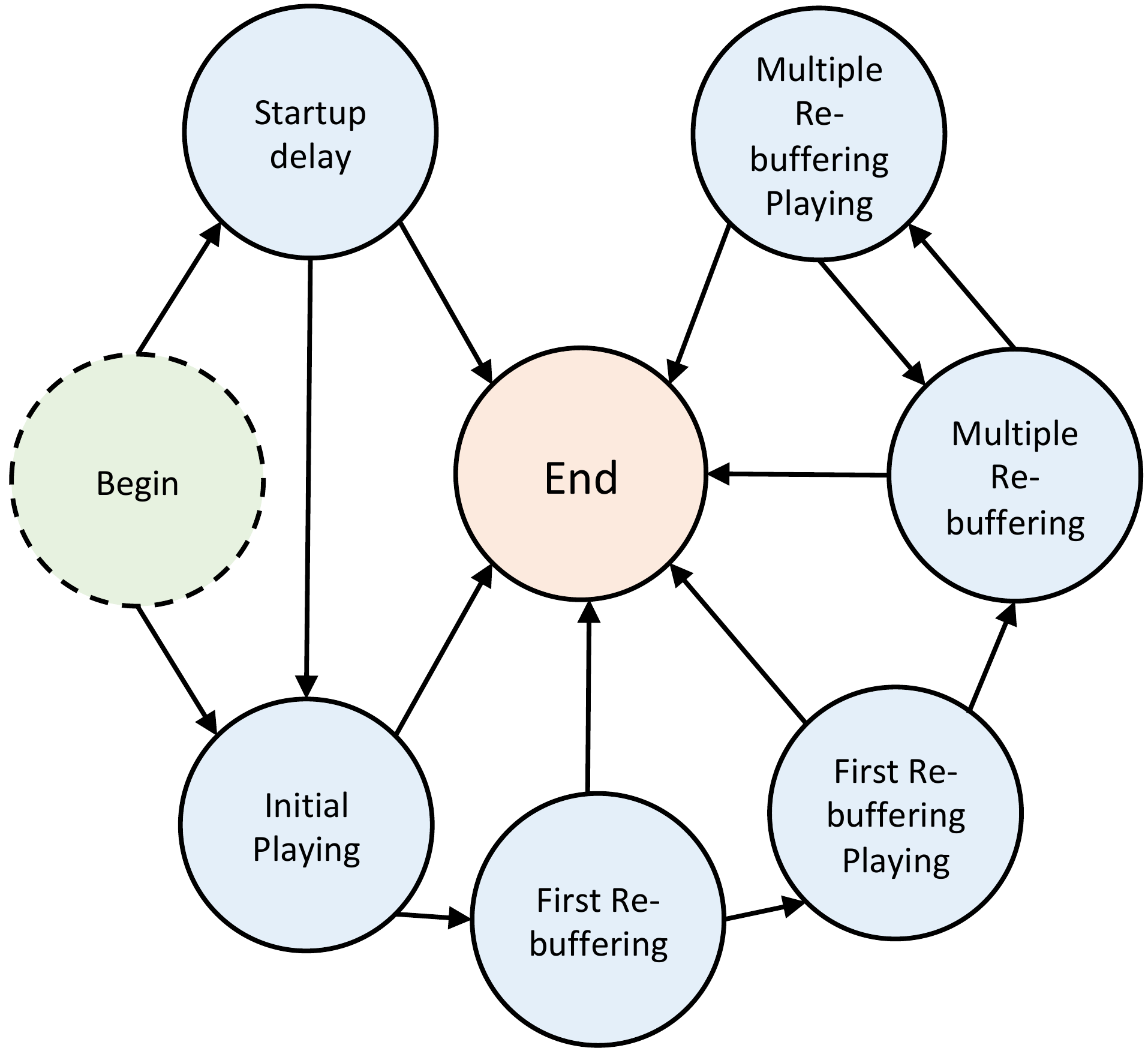}
\caption{DQS state machine}
\label{fig:dqs_model}
\end{minipage}
\hspace{-0.005in}
\begin{minipage}{.32\textwidth}
\centering
\includegraphics[width=0.9\columnwidth]{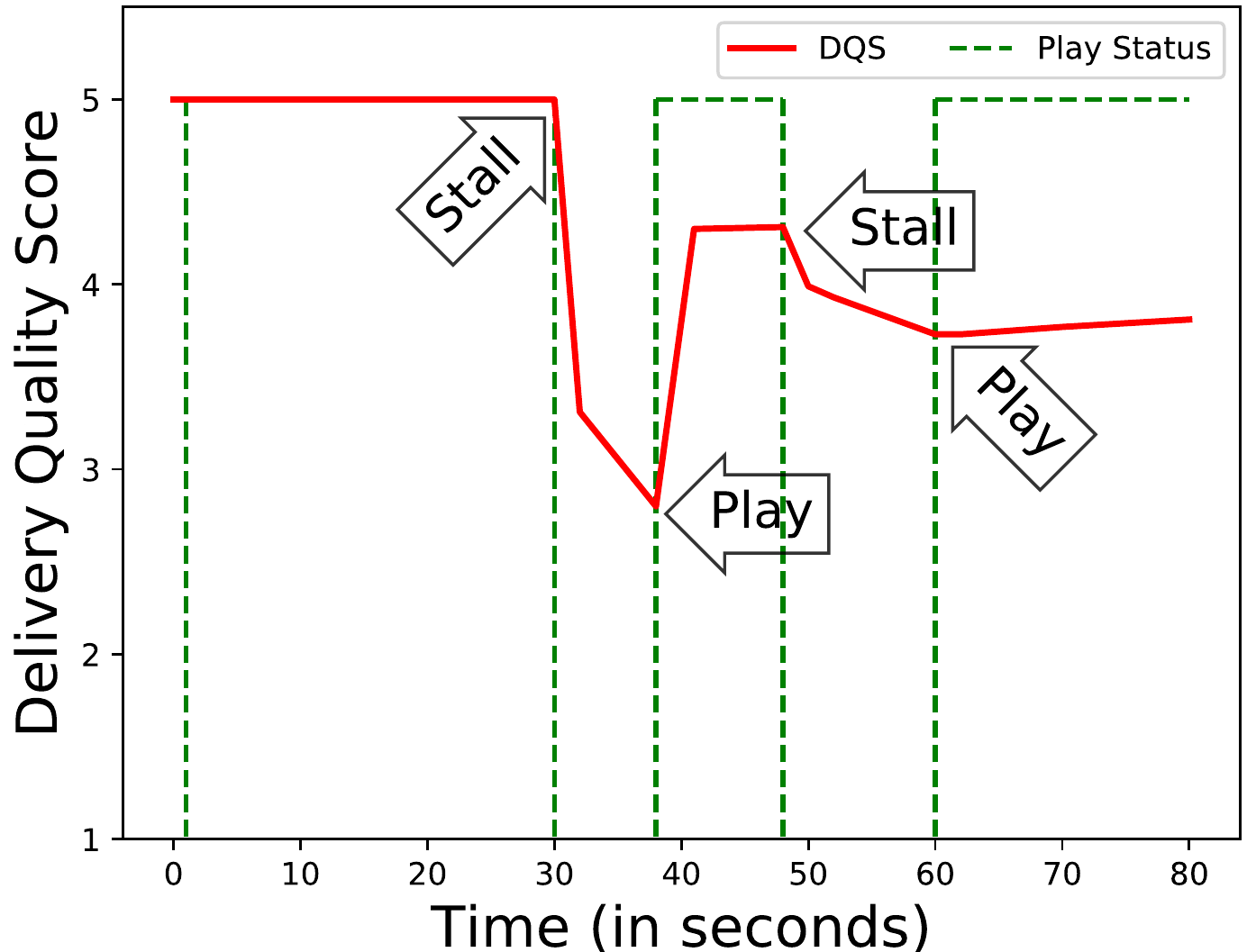}
\caption{Sample DQS evolution.}
\label{fig:dqs_evol}
\end{minipage}
\hspace{-0.05in}
\begin{minipage}{.32\textwidth}
\centering
\includegraphics[width=1.1\columnwidth]{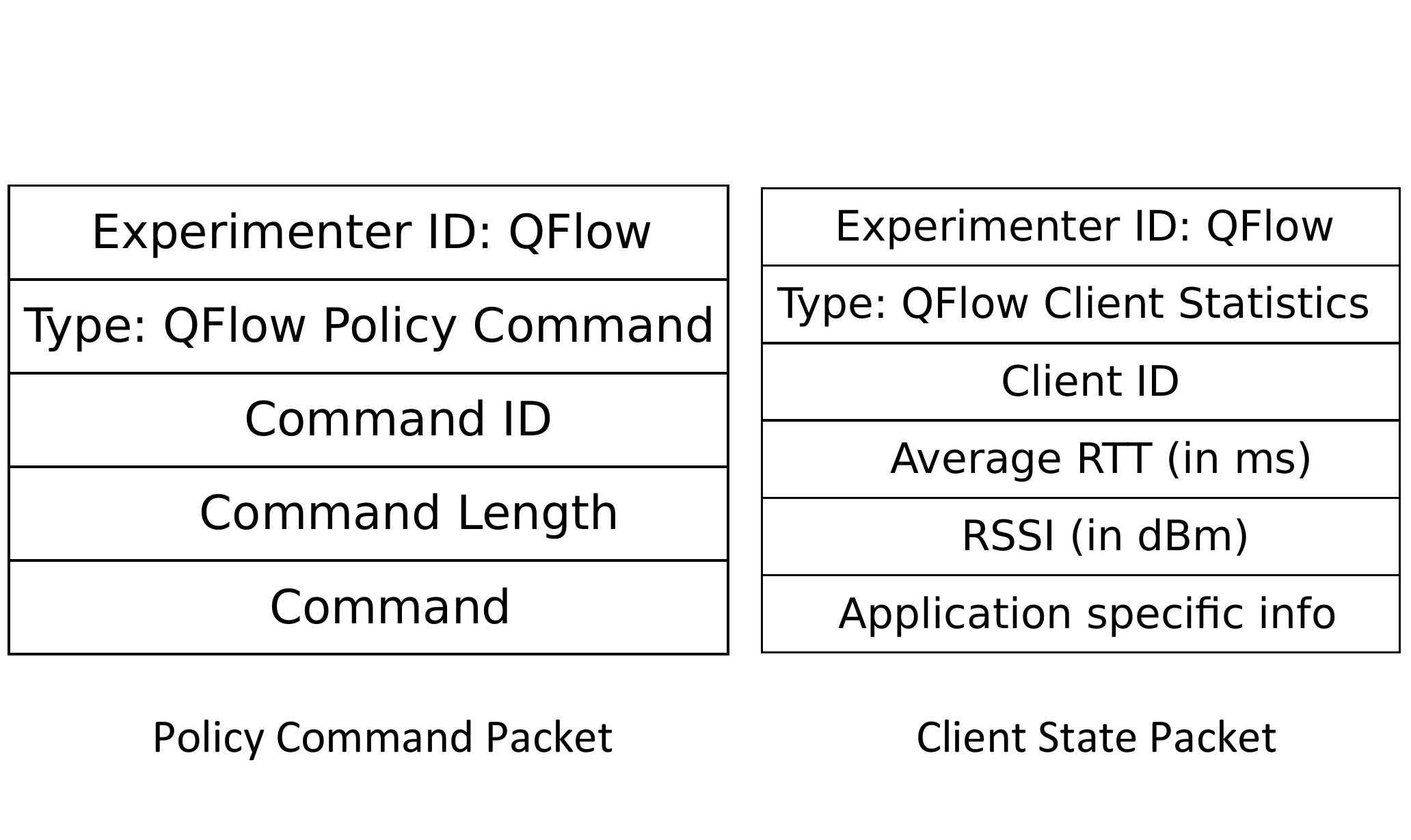}
\vspace{-0.15in}
\setcounter{figure}{5}
\caption{Packet formats in QFlow}
\label{fig:packets}
\end{minipage}
\vspace{-0.1in}
\end{figure*}
The DQS model weights the impact according to duration of the impairments to better model human perception.   
For example, the impact on QoE of stall events during playback is greater than that of initial buffering.  Similarly, the first stall event produces less dissatisfaction than repeated stalling.  
The state diagram of the model is shown in Figure \ref{fig:dqs_model}.  The increases and the decreases in perceived QoE are captured by a combination of raised cosine and ramp functions.  This enables it to model greater or lesser changes in the perceived QoE according to the time it spends in a particular state.  The behavior of the predicted QoE by the model in the presence of a particular stalling pattern can be seen in Figure \ref{fig:dqs_evol}, where the two stall events result in degradation of QoE.   Recovery of QoE from each stall event becomes progressively harder.

\subsection{QFlow System Architecture}

The system architecture is illustrated in Figure~\ref{fig:architecture}.  The three main units are, (i) an off-the-shelf WiFi access point running the OpenWRT operating system, (ii) a centralized controller hosted on a Linux workstation, and (iii) multiple wireless stations.  We denote each software functionality with both a color and a circled number.  These functionalities pertain to \textcircled{1} queueing mechanisms, \textcircled{2} QoS policy (configuration selection), \textcircled{3} Reinforcement Learning, and \textcircled{4} End User Value and Auction, which we overview below.  Tying together the units are \textcircled{5} Databases at the Controller (to log all events), and at each station (that obtains a subset of the data for local decision making).  The final components are \textcircled{6} Network Interfaces and \textcircled{7} User Application, which are unaware of our system.  We refer to the user application as a client or session, which is composed of one or more flows that are treated identically.
\begin{figure}[htbp]
\vspace{-0.1in}
\begin{center}
\includegraphics[width=3.3in]{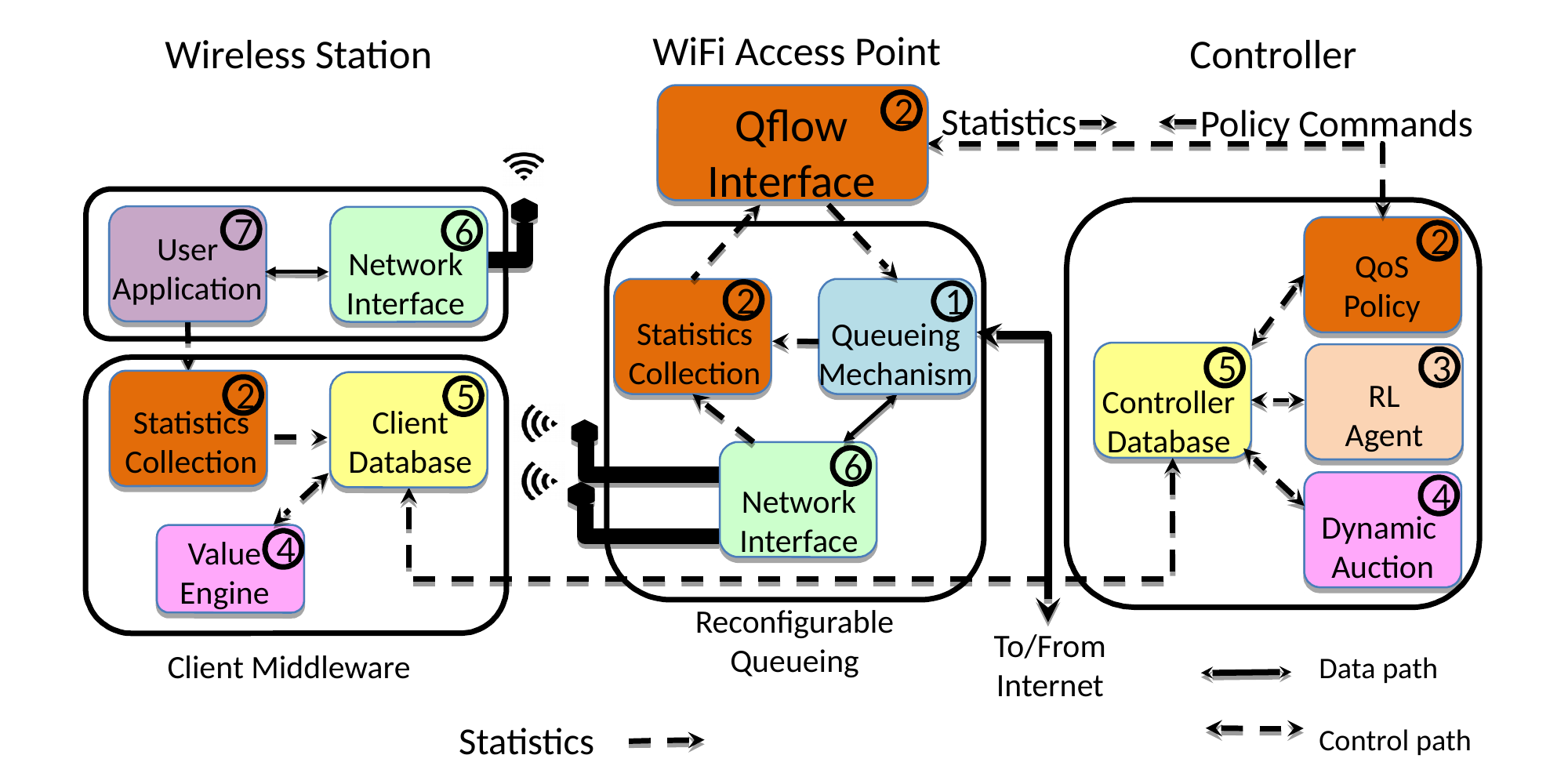}
\setcounter{figure}{4}
\caption{The system architecture of QFlow.
}
\label{fig:architecture}
\end{center}
\vspace{-0.15in}
\end{figure}

\textbf{\textcircled{1} Per-Packet Queueing Mechanisms:}
At the level of data packets, we utilize the MAC layer of software defined infrastructure, namely, reconfigurable queueing. Multiple Layer 2 queues can be created, and different per-packet scheduling mechanisms can be applied over them.  When such mechanisms are applied to aggregates of flows, the resulting QoS statistics at the queue level can be varied, with higher priority queues getting improved performances.  In turn, this results in state and QoE changes at the application.

\textbf{\textcircled{2} QoS Policy and Statistics:}
Policy decisions are used to select configurations (which clients are assigned to which queue).  Decisions are made at a centralized controller that communicates using the OpenFlow protocol.  We create a custom set of OpenFlow messages for QFlow.   The Access Point runs QFlow, which interprets these messages and instantiates the queueing mechanisms and configurations selected by the controller. The access point periodically collects statistics related to QoS, including signal strengths, throughput, and RTT and returns those back to the controller (these statistics are used for  sanity checks). 

A smart middleware layer at clients is used to interface with QFlow in a manner that is transparent to the applications (such as YouTube) and the end-user. The middleware determines the foreground application, and samples the application to determine its state (stalls, and buffered seconds on YouTube).  QoE is calculated using the DQS model.  
The client middleware contacts the Controller Database to  periodically send the application state and QoE.

\textbf{\textcircled{3} Reinforcement Learning Agent:} Application state and configuration decisions (state-action pairs) are used to train RL agents.  in the case of the model-free approach, a simulation environment duplicating the QFlow setup is used for offline training, and online training continues on the actual system.  In the case of model-based RL, state-action pairs (resulting from various different policies) stored in the controller database are used for learning the model.

\textbf{\textcircled{4} End User Value and Auction:}
Clients are offered feasible QoS vectors under an market framework.  The decision on which $N$ flows to admit to a high-priority queue is taken via an $N+1^{th}$ price auction using a local currency (a token allowance), which is conducted every $10$ seconds.  The resultant policy decisions in turn lead to a realization of the offered QoS.   End-users setup priorities for different applications (at the timescale of weeks or months), and the Controller Database provides statistics of current market conditions (bid distribution), using which a Value Engine at the client middleware determines what the value of winning and losing would be.  Finally, a Bid Generator places a bid for service.  Auction results translate into QoS policies that remain in place for $10$ seconds.

Policy Adaptation has to do with implementing the policy as empirical data accumulates.  An assignment algorithm (policy) matches sessions to queues every $10$ seconds, and obtains a sample of client state each time it does so.  This state-action pair is captured in the database, and a new action is obtained form the database (as determined by the agent).  The assignment algorithm is geared towards discounted QoE maximization.

\textbf{Interactions:}
The Client Middleware at each wireless client captures the state and calculates the corresponding QoE values specific to the foreground application.  These realized QoE and state values from all participating clients are sent to the Controller, which performs a policy decision for flow assignment.   These policy decisions are sent to the Access Point using OpenFlow Experimenter messages.  QFlow interprets and implements these policy decisions.  These steps are executed once every 10 seconds.  


\section{QFlow Implementation Details}
\label{section:softstack}
Our implementation of QFlow extends the OpenFlow protocol using experimenter messages.  We exploit the separation of control and data planes of OpenFlow to implement policy decisions using QFlow.  Further, our choice of using experimenter messages ensures that we do not require changes at the controller.  We use an off-the-shelf TP-Link WR1043ND v3 router with OpenWRT Chaos Calmer as the firmware for our implementation.  We choose OpenWRT because of its support for Linux based utilities like {\tt tc} (Traffic Control) for implementing per packet mechanisms. Since OpenWRT does not natively support SDN, we use CPqD SoftSwitch~\cite{cpqd}, an OpenFlow 1.3 compatible user-space software switch implementation.

We next extend SoftSwitch to include QFlow capabilities.  Such capabilities include the ability to modify packet-handling mechanisms.  Our goal is to enable configuration changes, in addition to the collection of statistics related to the implemented per packet mechanisms and the connected clients.  We construct two types of QFlow commands for implementing the described capabilities, \emph{Policy commands} and \emph{Statistics commands}. The rationale behind this separation is to differentiate policy decisions from statistics collection. The controller uses Experimenter messages to communicate these commands to the Access Point using OpenFlow.

\subsection{Policy Commands}
We design Policy commands to allow us to choose between available mechanisms at different layers.  Every time a Policy command is sent, it is paired with a \emph{Solicited response}  that is generated by the receiver
and sent to the controller using an experimenter message.  A Solicited response message thus provides us with feedback from the intended receiver.  
We define the format of the policy experimenter messages as shown in Figure~\ref{fig:packets} (left).  The Controller packs a policy command, and sends it to the Access Point using OpenFlow. On receiving the message, QFlow unpacks it, identifies the specific policy command using the type field, and performs the corresponding operation.  Using this framework, we implemented policy commands for the MAC layer.

\textbf{Data Link Layer Queue Command:}  At the data link layer, we need a means of providing variable queueing schemes.  Traffic control ({\tt tc}) is a Linux utility that enables us to configure the settings of the kernel packet scheduler by allowing us to \emph{Shape} (control the rate of transmission and smooth out bursts) and \emph{Schedule} (prioritize) traffic.  Each network interface is associated with a \emph{qdisc} (Queueing discipline) which receives packets destined for the interface.  We selected \emph{Hierarchical Token Bucket} ({\tt htb}) for our experiments because of the versatility of the scheme.  It performs shaping by specifying \emph{rate} (guaranteed bandwith) and \emph{ceil} (maximum bandwidth) for a class, with sharing of available bandwidth between children of the same parent class, and can also prioritize classes. Finally, we use \emph{Filter}s to classify and enqueue packets into classes.  

In our experiments, we create queues with different token rates using {\tt htb}.  Tokens may be borrowed between queues, meaning that queues will share tokens if they have no traffic.  We also create a default queue that handles any background traffic.
Decisions at the data link layer include assigning flows to queues, setting admission limits, changing the throughput caps queues, and enabling or disabling sharing of excess (unused) throughput between them.



\subsection{Statistics Commands} Policy commands result in changes to the QoS statistics of the queues. We define Statistics commands to collect these results and send them back to the controller for analysis. Queue statistics include cumulative counts of downlink packets, bytes and dropped packets. Client-specific statistics consist of average Round Trip Times (RTT; which includes both the RTT from the base station to the client as well as the RTT from the base station to the wide-area destinations with which the client communicates), signal strength (RSSI) and Application specific statistics like buffer state, stall information and video bitrate. Since statistics are sent periodically (once every second) to the controller, we label such messages as \emph{Unsolicited response messages}.

Similar to Policy commands, we define the structure of both Queue and Client-specific Statistics messages. After collecting the respective statistics, QFlow packs the data and sends them to the Controller using OpenFlow. On receiving these messages, the Controller unpacks them, identifies the type from the header information and then saves the extracted data to the database. The packet formats of the Client Statistics messages is shown in Figure~\ref{fig:packets} (right).  QFlow thus is capable of generating state-action, and measuring the resultant rewards in terms of QoE.    The details of using the system for RL will be described in the next two sections.


\section{Model-Free RL} 
\label{section:ModelFreeRL}

We describe a model-free RL based approach for learning a control algorithm for the system described in Section \ref{section:system}. More specifically, the objective is to learn a control policy  for the MDP  when the system model (transition probability kernel of the MDP) is unknown.  Model-free RL algorithms learn the optimal control policy directly via the interactions with the system, without explicitly estimating the system model. The interaction of the RL agent with the system is modeled as a set of tuples $\left(s_t, a_t, R_{t+1}, s_{t+1} \right)$ over time and the goal of the RL agent is to learn a policy $\pi$ that recommends an action to take given a state, in order to maximize its long term expected cumulative reward. We will employ a specific model-free RL algorithm known as Q-learning algorithm.

\subsection{Q-Learning}
Each state-action pair $\left(s,a\right)$ under a policy $\pi$ can be mapped to a scalar value, using a Q-function.  $Q^{\pi}(s,a)$ is the expected cumulative reward of taking an action $a$ in a state $s$ and following the policy $\pi$ from there on. $Q^{\pi}$ is specified as 
\begin{equation*}
Q^{\pi}(s,a) = \mathbb{E}\left[ \sum_{t=0}^{\infty} \gamma^t R(s_t,\pi(s_t))|s_0 = s, a_0 = a \right],
\end{equation*}
where $\gamma \in (0,1)$ is the discount factor.  Maximizing the cumulative reward is equivalent to finding a policy that maximizes the Q-function.  The optimal  Q-function, $Q^{*}$ satisfies the Bellman equation
\begin{equation*}
Q^{*}(s,a) = R(s, a) + \gamma \mathbb{E}_{s'}[ \max_{b} Q^{*}(s',b)], \forall s, a.
\end{equation*} 
The objective of the Q-learning algorithm is to learn this optimal $Q^{*}$ from the sequence of observations $\left(s_t, a_t, R_{t+1}, s_{t+1} \right)$. The optimal policy $\pi^{*}$ can be computed from $Q^{*}$ as,
\[\pi^{*}(s)  = \arg \max_{a} Q^{*}(s, a).  \] 

The Q-learning algorithm is implemented as follows. At each time step $k$, the RL agent updates the Q-function  $Q_{k}$ as  
\begin{equation*}
Q_{k+1}(s,a) = 
\begin{cases}
     (1-\alpha_k)Q_k(s,a)+ \alpha_k ( R_k \\ \hspace{10pt}+\ \gamma \max_b Q_k(s_{k+1},b)) 
          \hspace{10pt} \text{if $s= s_k, a = a_k$}\\
     Q_k(s,a)    \hspace{83pt} \text{otherwise}
\end{cases} 
\end{equation*} 
where $\alpha_{k}$ is the learning rate. If each-state action pairs is sampled infinitely often and under some suitable conditions on the step size, $Q_{k}$ will converge to the optimal Q-function $Q^{*}$ \cite{sutton2018reinforcement}.


\subsection{Deep Q-Learning}
Using a standard tabular Q-learning algorithm as described above to solve our problem is  infeasible  due to the huge state space associated with it.  The individual client states are combined to form a joint state. The aggregate reward is the reward of all clients combined. The learning agent observes the states and rewards, and outputs an action. The environment then moves to the next state, yielding a reward. 


The state of each client is a tuple consisting of its buffer state, stall information, and its QoE at $t$. 
Buffer state and QoE are considered to be real numbers, and thus can take an uncountable number of values. Even if we quantize, the number of states increases exponentially with the dimension and the number of clients. Tabular Q-learning approaches fails in such scenarios.

To overcome this issue due to the curse of dimensionality, we address this problem through the framework of  deep reinforcement learning. In particular, we use the double DQN algorithm  \cite{van2016deep}. This approach is  a clever combination of three main ideas:  Q-function approximation with neural network, experience replay, and target network. We give  a brief description below. 

 \textbf{Q-function approximation with neural network:}
To address the problem of large and continuous  state space, we approximate the Q-function using a multi-layer neural network, i.e., $Q(s,a) \approx Q_{w}(s,a)$ where $w$ is the parameter of the neural network. Deep neural networks have achieved tremendous success in both supervised learning (image recognition, speech processing) and reinforcement learning (AlphaGo games) tasks.  They can approximate arbitrary functions without explicitly designing the features like in classical approximation techniques. The parameter of the neural network can be updated using a (stochastic) gradient descent with step size $\alpha$ as
\begin{equation}
\label{eqn:sgd}
w = w + \alpha \nabla Q_w(s_t,a_t) (R_t + \gamma \max_b Q_w(s_{t+1},b) - Q_w(s_t,a_t))
\end{equation}

%

 \textbf{Experience Replay:}
Unlike supervised learning algorithm, the data samples $\{s_t, a_t, R_t, s_{t+1}\}$ obtained by an RL algorithm is correlated 
in time due to the underlying system dynamics. This often leads to a very slow convergence or non-convergence of the gradient descent algorithms like \eqref{eqn:sgd}. The idea of experience replay is to break this temporal correlation by randomly sampling some data points from a buffer of previously observed (experienced) data points  to perform the gradient step in \eqref{eqn:sgd} \cite{mnih2015human}. New observation are then added to the replay buffer and the process is repeated. 

 \textbf{Target Network:}
In  \eqref{eqn:sgd}, the target  $R_t + \gamma \max_b Q_w(s_{t+1},b)$ depends on the neural network parameter $w$, unlike the targets used for supervised learning which are fixed before learning begins. This often leads to poor convergence in RL algorithms. To addresses this issue, deep RL algorithms maintain  a separate neural network for the target. The target network is kept fixed for multiple steps. The update equation with target network is given below.
\begin{align*}
	&w = w + \alpha \nabla Q_w(s_t,a_t) (R_t + \gamma \max_b Q_{w^-}(s_{t+1},b) \\ & \hspace{180pt}- Q_w(s_t,a_t))\\
 &w^- = w \quad \text{after every $N$ steps.}
\end{align*}

The combination of neural networks, experience replay and target network forms the core of the DQN algorithm \cite{mnih2015human}. However, it is known that DQN algorithm suffers from overestimation of Q values.  Double DQN algorithm \cite{van2016deep} overcomes this problem using slightly modified updated equation as
\begin{align*}
	w = w + \alpha \nabla Q_w(s_t,a_t) (R_t + \gamma Q_{w^-}(s_{t+1},& \\ \argmax_b Q_w(s_{t+1},b)) - Q_w(s_t,a_t)).
\end{align*}
The target network is updated after every $N$ steps as before.


\subsection{Training the RL Algorithm}
We implemented the double DQN algorithm using the TensorForce library \cite{schaarschmidt2017tensorforce}. Hyperparameters are selected via random search. The final configuration and hyperparameter of the RL algorithm is specified in  Table \ref{tab:hyperparameters}.

\begin{table}
\begin{center}
\caption[RL Hyperparameter Selection]{Selected hyperparameters for RL agent}
  \label{tab:hyperparameters}
\begin{adjustbox}{max width=\linewidth}
  \begin{tabular}{|l|r|}
    \hline
    \emph{Hyperparameter}   & \emph{Chosen Value}	        												\\ \hline
    Discount                & $0.9999$	        															\\ \hline
    Network Hidden Layers   & $(64, 32)$        															\\ \hline
    Network Optimizer       & Adam, Learning Rate $0.001$													\\ \hline
    Replay Buffer           & $500000$           															\\ \hline
    Replay Batch            & $32$              															\\ \hline
    Target Sync Period      & $100000$          															\\ \hline
    Huber Loss              & $1.0$             															\\ \hline
    Double Learning         & On                															\\ \hline
    Control Policy          & $\epsilon$-greedy, Decay $\epsilon$ from $1.0$ to $0.01$ over $1000000$ steps	\\
    \hline
  \end{tabular}
  \end{adjustbox}
\end{center}
\end{table}


For the faster training of our RL algorithm, we first implement a simulation environment which closely mimics the dynamics of the physical testbed. The environment simulates each video including its bitrate, buffer, length, and QoE. The bitrate and length of each video is generated according to a normal distribution; buffer is stored in terms of time, rather than bits. Each client continuously plays one video after another, stalling where its buffer runs out and building up a buffer of 10 seconds before attempting playing again. Queues are serviced with a constant total bandwith, but the fairness of queue's service among flows assigned to that queue is chosen in each decision period (DP) according to a Dirichlet distribution. Each DP is of duration $10$ seconds.  The simulation environment uses a high-priority queue with 11 Mbps bandwidth and a low-priority queue with 4.3 Mbps. In the static network configuration, six clients are specified that draw video bit-rates from a truncated $\mathcal{N}(2.9, 10)$ distribution in Mbps, and draw video lengths from a truncated $\mathcal{N}(600, 50)$ distribution in seconds.

\begin{figure}[htbp]
\begin{center}
\vspace{-0.1in}
\includegraphics[width=3.4in, height=2.3in]{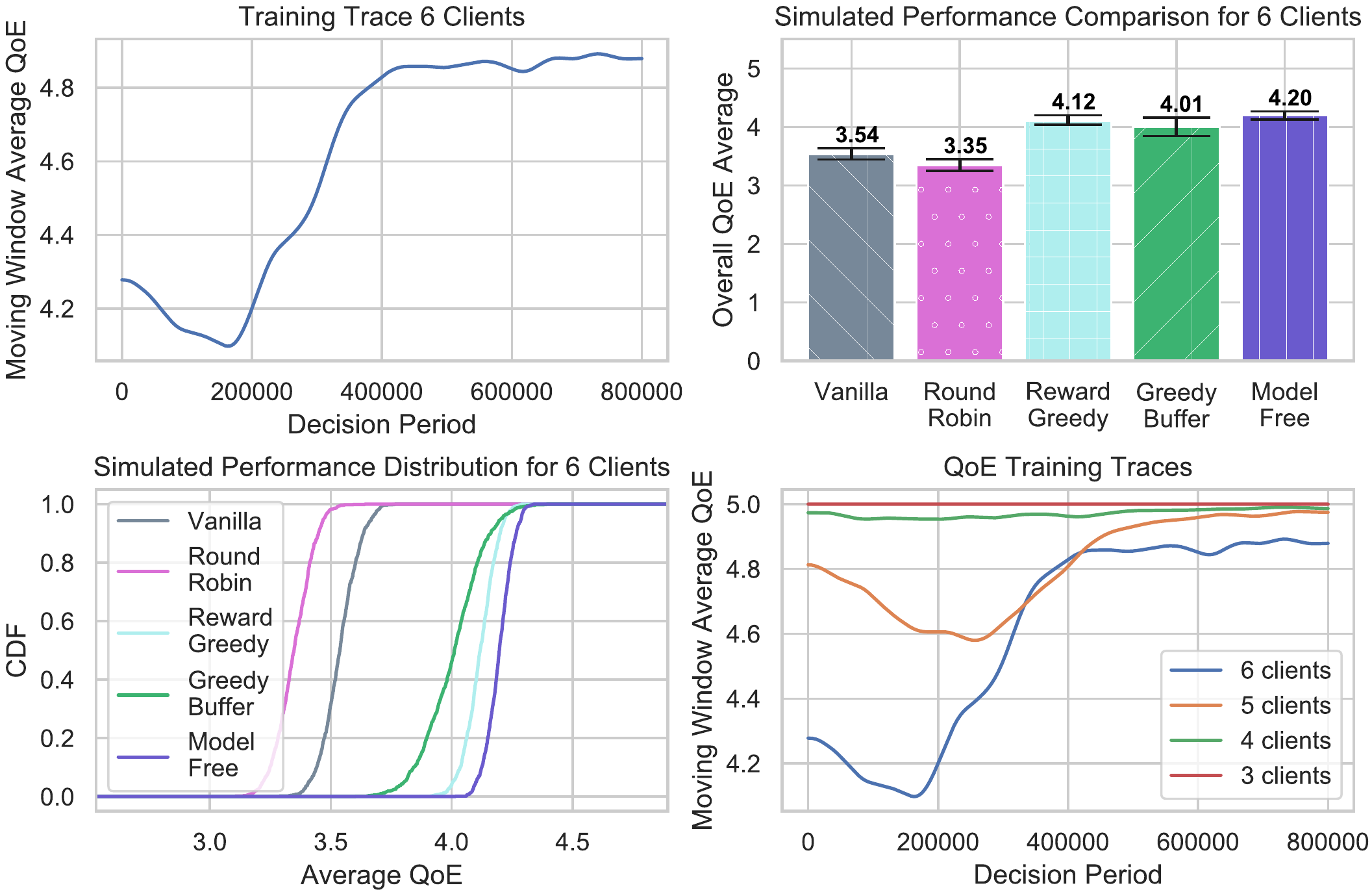}
\setcounter{figure}{6}
\caption{Training model-free RL via simulations}
\label{fig:training}
\end{center}
\vspace{-0.1in}
\end{figure}


For hyperparameter search, the system was simulated for 200 DP per episode for 1000 episodes. Note that increasing the number of units or layers in the network used for value estimation after $(64,32)$ does not significantly affect the convergence curve; however, the magnitude of the learning rate creates large differences in the performance to which the agent ultimately converges. Further, a single layer is incapable of learning to the performance achieved by the two-layer network. We therefore choose the $(64,32)$ configuration for our agent.   The evolution of value during the training process is shown in Figure~\ref{fig:training} top-left.  As is seen, the trained controller achieves a high QoE of near 5.

Next, we compare the performance of different policies in the simulation environment.  Figure~\ref{fig:training} top-right shows the average QoE attained by different policies, which suggests that perhaps the model-free approach, while best, may not give substantial performance improvements.   The QoE CDFs in Figure~\ref{fig:training} bottom-left, however, indicate that model-free RL achieves a higher QoE for a larger fraction of clients, suggesting that it might be more robust to resource constraints.  Indeed, we will see in experiments in Section~\ref{section: experiments} that it attains quite substantial gains over the other approaches in practice under a bandwidth constrained environment.

\subsection{Dynamic Number of Clients}
In the above description, we assumed that the number of clients in the system is static.   The timescale at which the number of clients change is very large (several tens of minutes; this models human mobility) when compared to the decision period (10 seconds).  Including a dynamic number of clients into training would require augmenting the state space with the number of connected clients, and a Markov model of transitions in this value.   Since this increases the state space and training duration, we instead obtain the optimal static policy for the system with 4 to 6 clients using the model-free approach.  Figure~\ref{fig:training} bottom-right shows the evolution of value over the training process over the different cases.  We can then choose the the right policy based on number of clients in the system.    Interestingly, there appears to be enough structure in our problem that a policy developed for a larger number of clients can simply be used for a smaller number (setting non-existent clients to have large QoE and buffer values), since the relative priorities of clients is all that matters.  We discuss this idea further in Section~\ref{section:index-policy}.

\section{Model-Based RL}
\label{section:ModelBasedRL}
In this section, we discuss the scenario in which the dynamics of the system (transition kernel) are first determined, i.e., given the current state $s_t$ of the system and the action taken $a_t$, we find the transition probabilities to the next states $s_{t+1}$.  Given the transition kernel of the system $P$, we can use  policy or value iteration to solve for the optimal policy $\pi^{*}$.  The model-based approach is particularly interesting because of its special structure, since the state transitions of a client given its current state and action are independent of the states and actions of other clients in the system.  In other words,  
\begin{equation}
\label{eqn:product-form}
{P}(s_{t+1}|a_t,s_t)=\prod_{\forall c \in \mathcal{C}}{P}(s^c_{t+1}|a_t^c,s^{c}_t)
\end{equation}
It must also be noted that the state transitions of all clients in the system given their current states and actions are identical.  Thus, we can determine the transition kernel of the system using the transition kernel of each individual client.
\subsection{Static Model}
 In what follows, we determine the transition kernel of the system with a fixed number of clients, and obtain the optimal policy.
  
\textbf{Experimental Traces.} We generate state ($s_t^c$), action ($a_t^c$) and next state ($s_{t+1}^c$) tuples for all  clients by running the system under Round Robin, Greedy Buffer, Random, Model Free and Vanilla policy  for a duration of 10 hours. 

\textbf{Discretizing the state space.} The state of each individual client $s^c_t$ and hence the state of the system $s_t$ have elements that are (non-negative) real numbers.  In order to calculate the transition kernel of the client in  atractable manner, we discretize the state space of the client according to table \ref{Discretized parameters}.  Since the state of a client is 3 dimensional (Buffer, Stall, QoE) we encode it to obtain a label for each client state as follows, Let $NSB$ and $NQB$ denote the number of stall and QoE bins respectively,  
$$s_t^c = \text{Buffer}\times \text{NSB} \times \text{NQB } + \text{QoE}\times \text{NSB } + \text{Stall} $$
The discretized and encoded state space of a client $\mathcal{S}_c$ has a cardinality of $945$. 

\textbf{Determining the transition kernel of a client.} We determine the transition kernel of a single client by fitting an empirical distribution over the state, action, and next state tuples collected from experimental traces, i.e., we empirical determine,  
$${P}(s^c_{t+1}|a_t^c,s^{c}_t) \quad \forall s^c_{t+1}, s^c_{t} \in \mathcal{S}_c  \quad \forall a^c_t \in \mathcal{A}_c $$ 
from experimental traces.  $\mathcal{A}_c$ is the set of all actions for a client $c.$

\textbf{Identifying Frequent States of the system.} The state of the system ($s_t$) is the union of states of all clients ($s_t^c$) in the system.  If there are $N$ clients in the system, the state of the system is a $N$ dimensional vector,   where each dimension corresponds to the state of a client.  Let $\mathcal{S}$ denote the discretized state space of the system. The cardinality of $\mathcal{S}$ is of the order of $945^N$.  Solving an MDP with $945^N$ states is intractable.  Hence, based on experimental traces, we identify the most frequent states $\mathcal{S}_p$ of the system, and approximate all other states to these popular states using the $L^2$ norm, i.e., given a state in $\mathcal{S}$, we approximate it by a state in $\mathcal{S}_p$ with the least Euclidean distance.   

\textbf{Calculating the transition kernel of the system.} The state space of our system has now reduced from $\mathcal{S}$ to $\mathcal{S}_p$.  To obtain the transition kernel of this system, we empirically sample one hundred state transitions for  each state in $\mathcal{S}_p$ under each action in $\mathcal{A}$ using the transition kernel of individual clients.  If the generated state transitions are outside  $\mathcal{S}_p$, we approximate it with the state in $\mathcal{S}_p$ which is closest in Euclidean distance. Thus, we obtain state, action, next state tuples for the system with state space $\mathcal{S}_p$.  We fit an empirical distribution over these tuples to obtain the transition kernel of the system.  Hence,  we empirically determine 
$$\tilde{P}(s_{t+1}|a_t,s_t) \quad \forall s_{t+1}, s_{t} \in \mathcal{S}_p  \quad \forall a_t \in \mathcal{A}.$$

\textbf{Obtaining the optimal policy} We obtain the optimal policy $\pi^*$ by running value iteration over the transition kernel generated for $\mathcal{S}_p$. It must be noted that the reward obtained by taking action $a_t$ in state $s_t$ is the average QoE of state $s_{t+1}$ which is a part of $s_{t+1}$ and hence need not be calculated explicitly. 

\begin{table}[]
\caption{Client State Space Discretization}
\label{Discretized parameters}
\centering
\begin{tabular}{|l|l|l|}
\cline{1-3}
\textbf{Parameter} & \textbf{Range} & \textbf{Bins}  \\ \cline{1-3}
Buffer             & {[}0,20{]}     & 21                \\ \cline{1-3}
Stalls             & {[}0,5{]}      & 5                \\ \cline{1-3}
QoE                & {[}1,5{]}      & 9                \\ \cline{1-3}
\end{tabular}
\vspace{-0.1in}
\end{table}
\subsection{Dynamic Number of Clients}
 In the previous subsection, we assumed that the number of clients in the system are static.  To deal with a dynamic number of clients, we follow an approach similar to the one described in section \ref{section:ModelFreeRL}.  We obtain the optimal policy for the system with 4-6 clients using the static model approach described in the previous subsection.  In the same manner as the model-free case, we may also use the policy for 6 clients for a smaller number of clients by just comparing their relative priorities, as will be discussed in Section~\ref{section:index-policy}.




\section{Auction}
\label{section:auction}
As discussed in the Introduction, the model-based and model-free approaches require the self-reporting of states.  We now consider the auction system, wherein agents place bids to determine their relative valuations for priority service.  To determine its value, a client must solve (\ref{eq:val}) and obtain the optimal value function.  The information required to compute this solution are the transition kernel and the bid (belief) distribution of agents.   This belief distribution is obtained from the auction server (Controller), which collects the bids made over intervals of time and provides the empirical distribution back to all agents.Furthermore, the model-based approach immediately provides the transition kernel using the transition kernel of a client (Section \ref{section:ModelBasedRL}).

The auction itself is chosen as an  $(N+1)^{th}$-price auction with $N$ identical goods (i.e., the number of clients that may be admitted to the high-priority queue) in which each agent may obtain at most one unit of the good.  It is straightforward to see that such an auction is a simple VCG auction \cite{krishna2009auction}, and so is incentive compatible (agents bid true values obtained from solving (\ref{eq:val})).   Hence, the approach provides a map between state of a client and its bid, which is subsequently used in the $(N+1)^{th}$ price  auction. The Auction Agent receives the bids from all clients, conducts the $(N+1)^{th}$-price auction and performs the assignment on the basis of the result.

\section{Index Policy and Dynamic Number of Clients}
\label{section:index-policy}
The auction approach suggests that at equilibrium each client is associated with a value that only depends on the state of that client, and the transition kernel (\ref{eqn:product-form}).  The solution to (\ref{eq:val}) results in a \textit{value} for each state  $s_t^{c}$ of the client.  Then in the manner of the Whittle Index~\cite{whittle1988restless}, we can order states in increasing order of value, and associate each state with an \emph{index,} which is its position in the order.    Then these indices can be used to directly decide which clients to prioritise, and we call this as an \emph{index policy} that simply picks the clients with the highest indices to provide priority service. 

Now, given the indices corresponding to a system with $J$ clients, it would save computational effort if we could use the same indices for a system with $K < J$ clients, by simply setting indices of non-existent clients to  0.  For example, would the indices for a system with 6 clients be consistent with one that has 3 clients?
\begin{figure}[htbp]
\vspace{-0.1in}
\begin{center}
\includegraphics[width=3in]{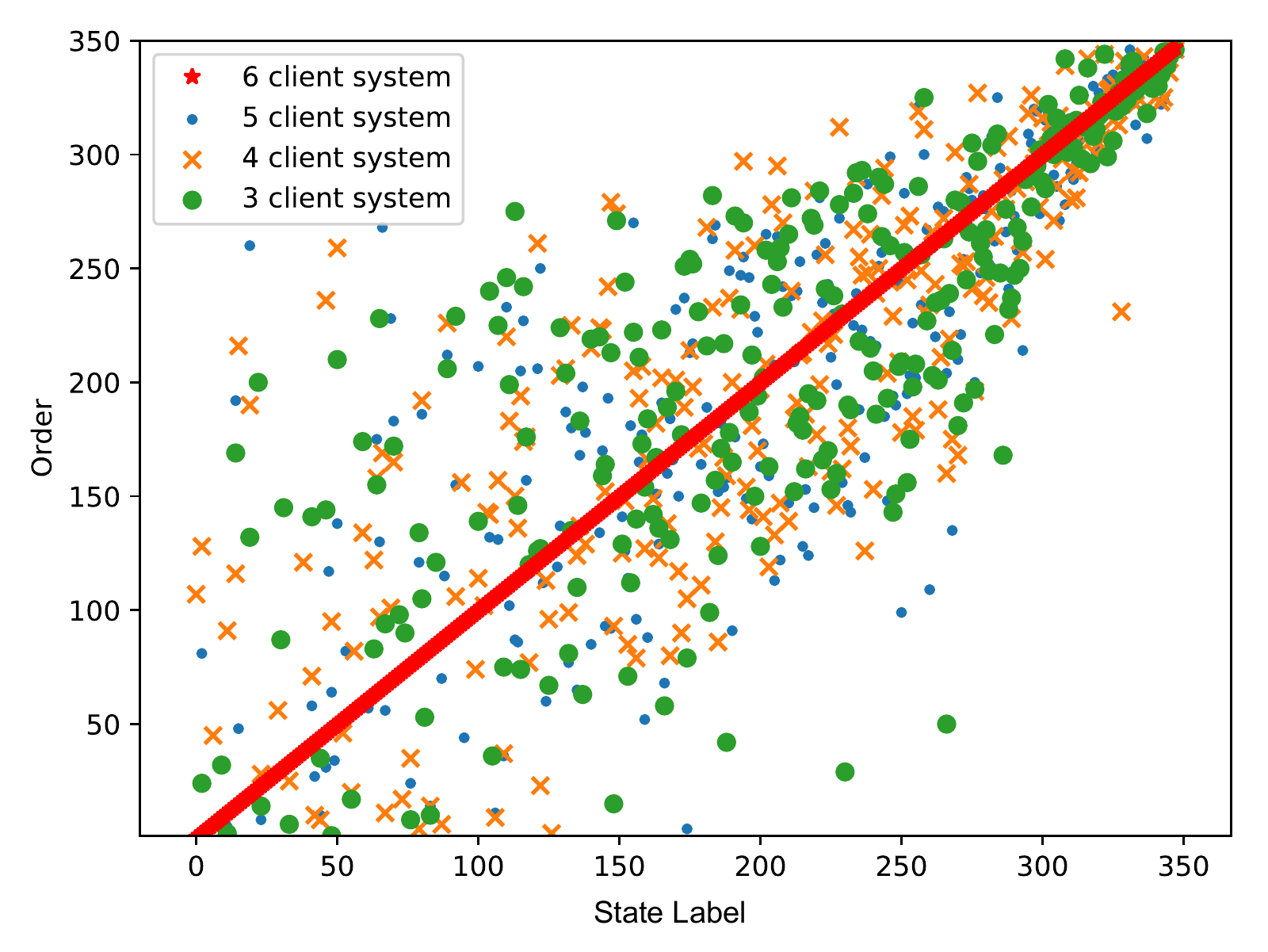}
\caption{Ordering of states in different client configurations.}
\label{fig:index}
\end{center}
\vspace{-0.2in}
\end{figure}


We experimentally determined the values for different numbers of clients, and determined the ordering of states in each case.   The comparison of the orderings for different client configurations (6, 5, 4, 3 clients) is shown in Figure~\ref{fig:index}, using the ordering for 6 clients as the base ordering.  We have not shown several hundred states that are indistinguishable at minimal value, and are all assigned an order of 0.  Specifically,  we find the values of the 350 states that turn out ot have non-minimal value with 6 clients, and assign the label 350 to the highest value state, 349 to next highest value state and so on.  Hence, when ordered by value, state 350 is also the 350$^{th}$ in order and so on, which results in the 45$^\degree$ red line in Figure~\ref{fig:index}.  Next, we find the values of each state for 5, 4 and 3 clients, and in each case show what is the ordering of the states using the same state labels as we did in the case of 6 clients.  We observe that the relative ordering of most of the high value states is consistent across configurations.

The above observation indicates that it is unnecessary to obtain policies tuned to the number of clients.  Rather, simply training the system with a fixed number of clients and using the relative state priorities obtained from such training for an instance with a smaller number of clients is likely to perform well.   In particular, a Whittle-index like policy developed for 6 clients is likely to prioritize the correct clients in a system with fewer clients.  This is the approach that we use in the next section while considering experiments with a dynamically changing number of clients.




\section{Evaluation}
\label{section: experiments}

\begin{figure*}[htbp]
\centering
\begin{minipage}{.32\textwidth}
\centering
\includegraphics[width=1\columnwidth]{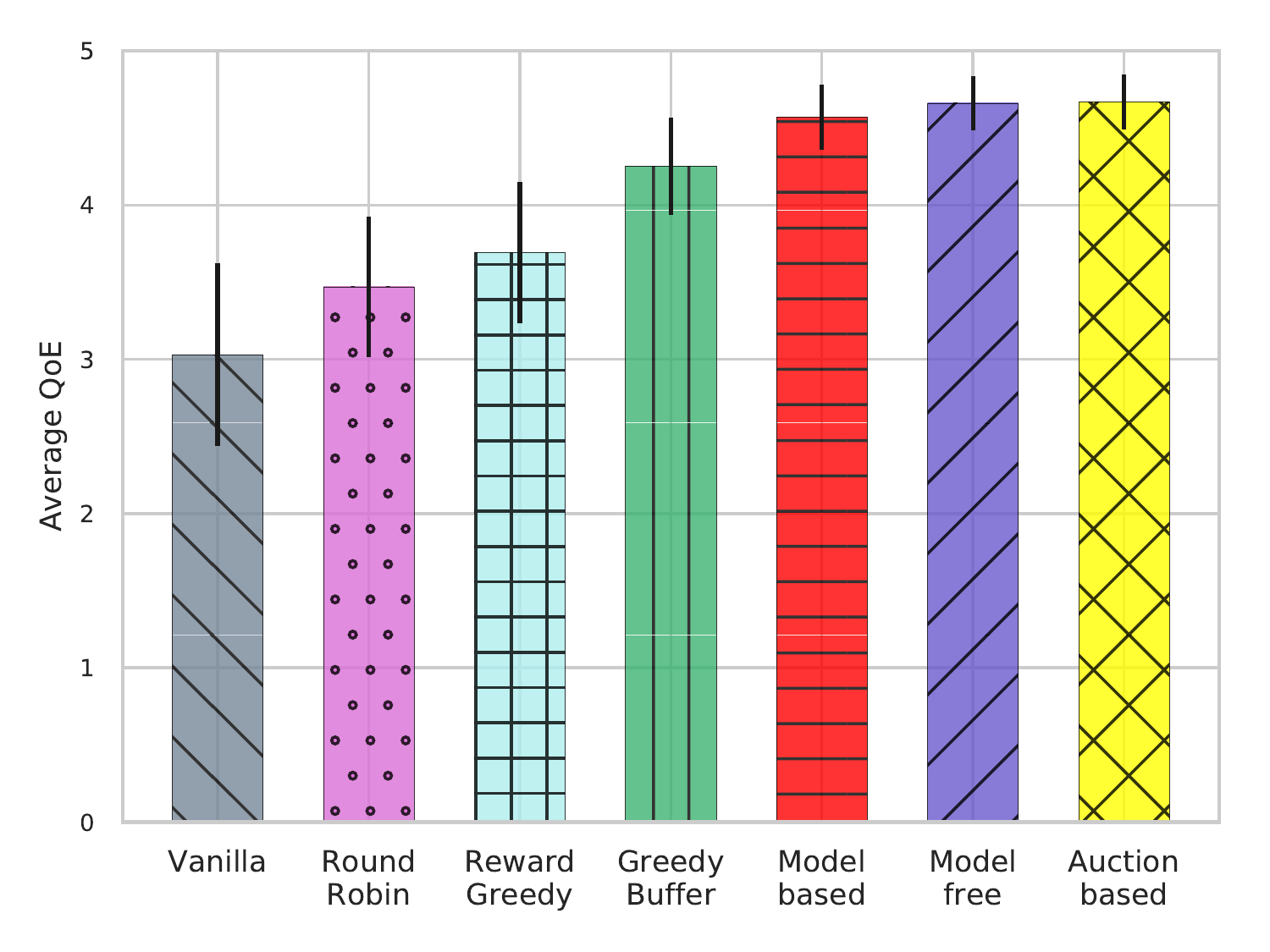}
\caption{Comparison of average QoE }
\label{fig:qoe_comp}
\end{minipage}\hfill
\begin{minipage}{.32\textwidth}
\centering
\includegraphics[width=1\columnwidth]{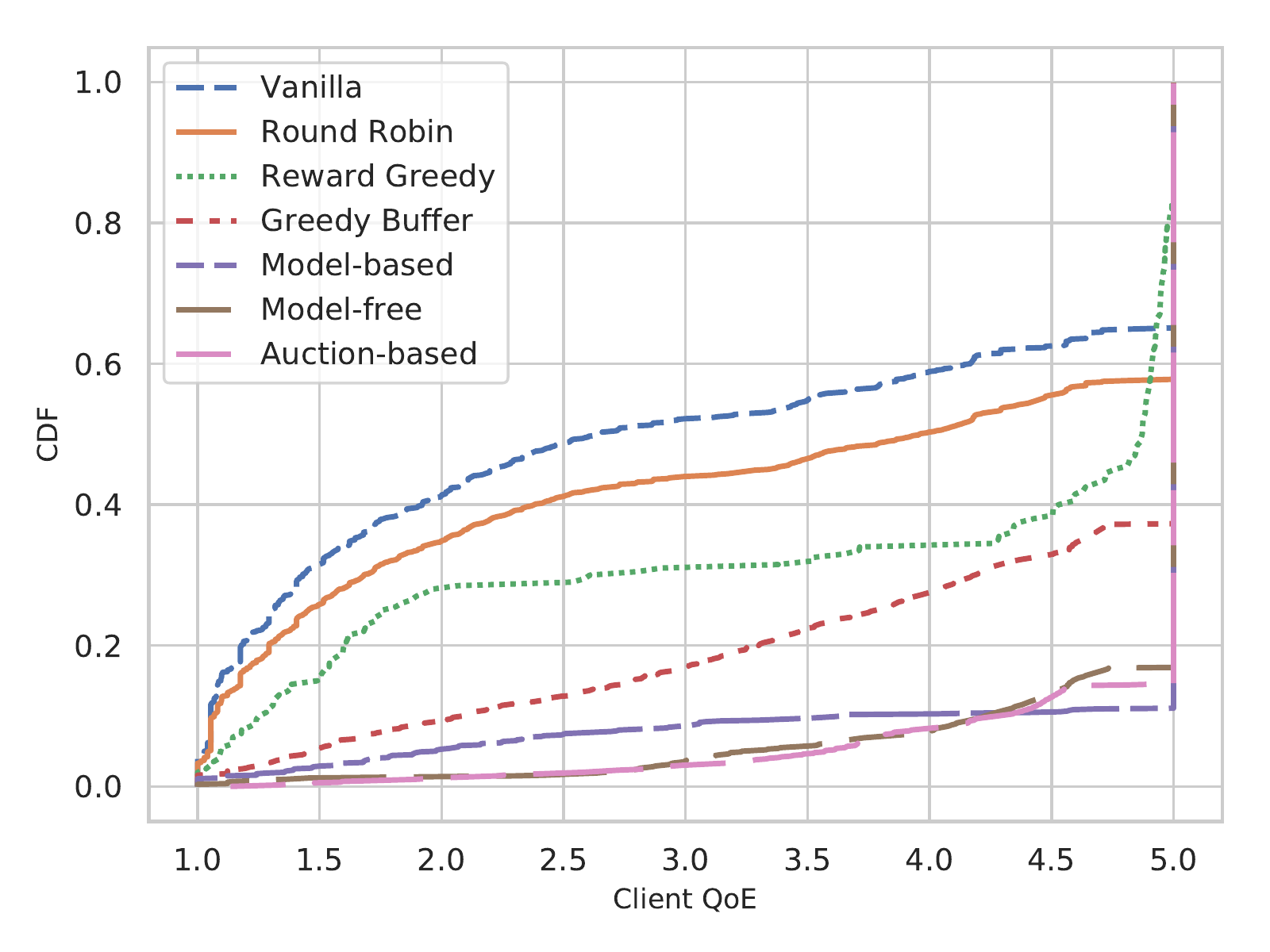}
\caption{Comparison of client QoE CDF }
\label{fig:qoe_cdf}
\end{minipage}\hfill
\begin{minipage}{.32\textwidth}
\centering
\includegraphics[width=1\columnwidth]{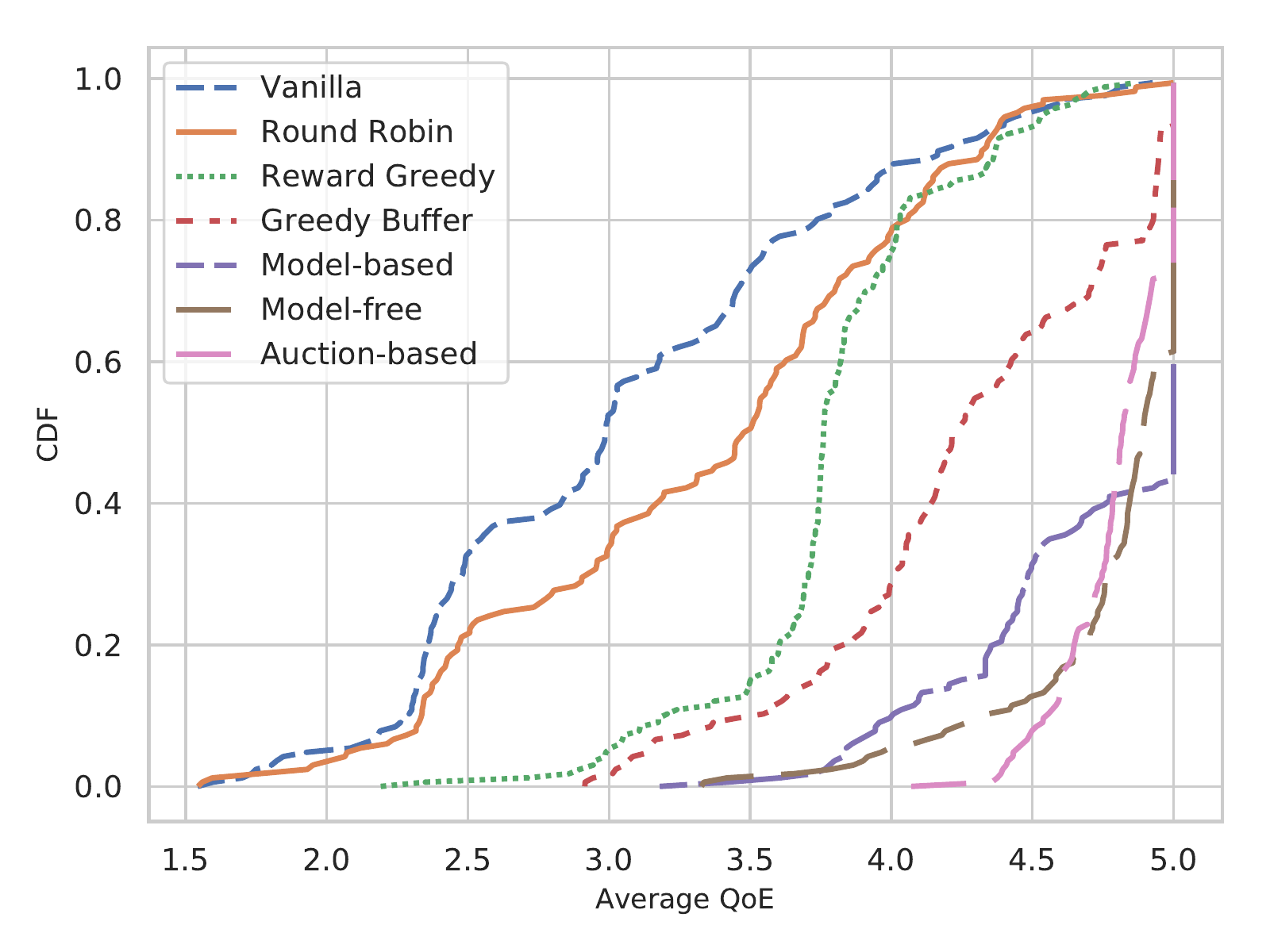}
\caption{Comparison of average QoE CDF }
\label{fig:qoe_avg_cdf}
\end{minipage}
\begin{minipage}{.32\textwidth}
\centering
\includegraphics[width=1\columnwidth]{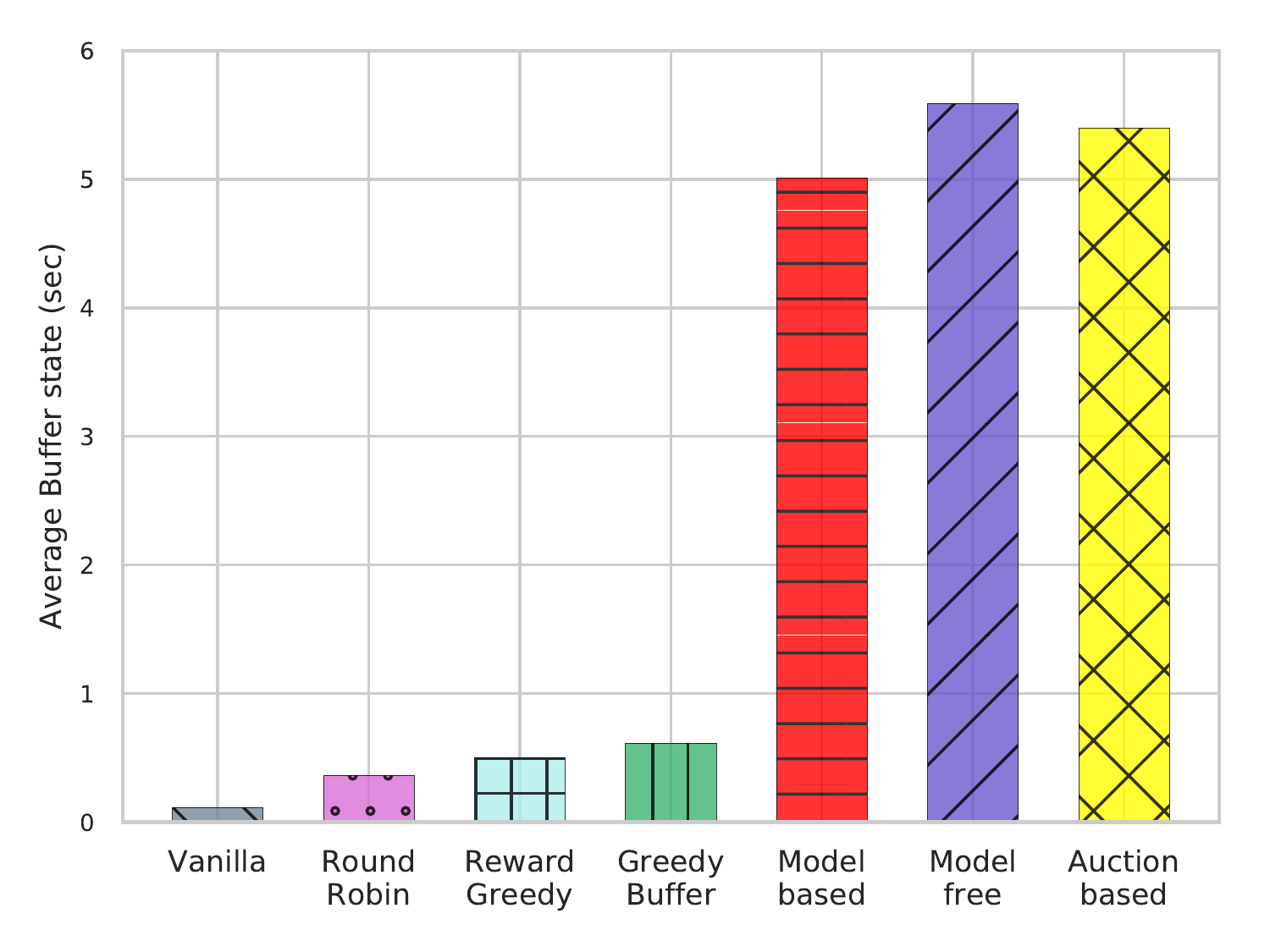}
\caption{Comparison of average Buffer }
\label{fig:buf_comp}
\end{minipage}\hfill
\begin{minipage}{.32\textwidth}
\centering
\includegraphics[width=1\columnwidth]{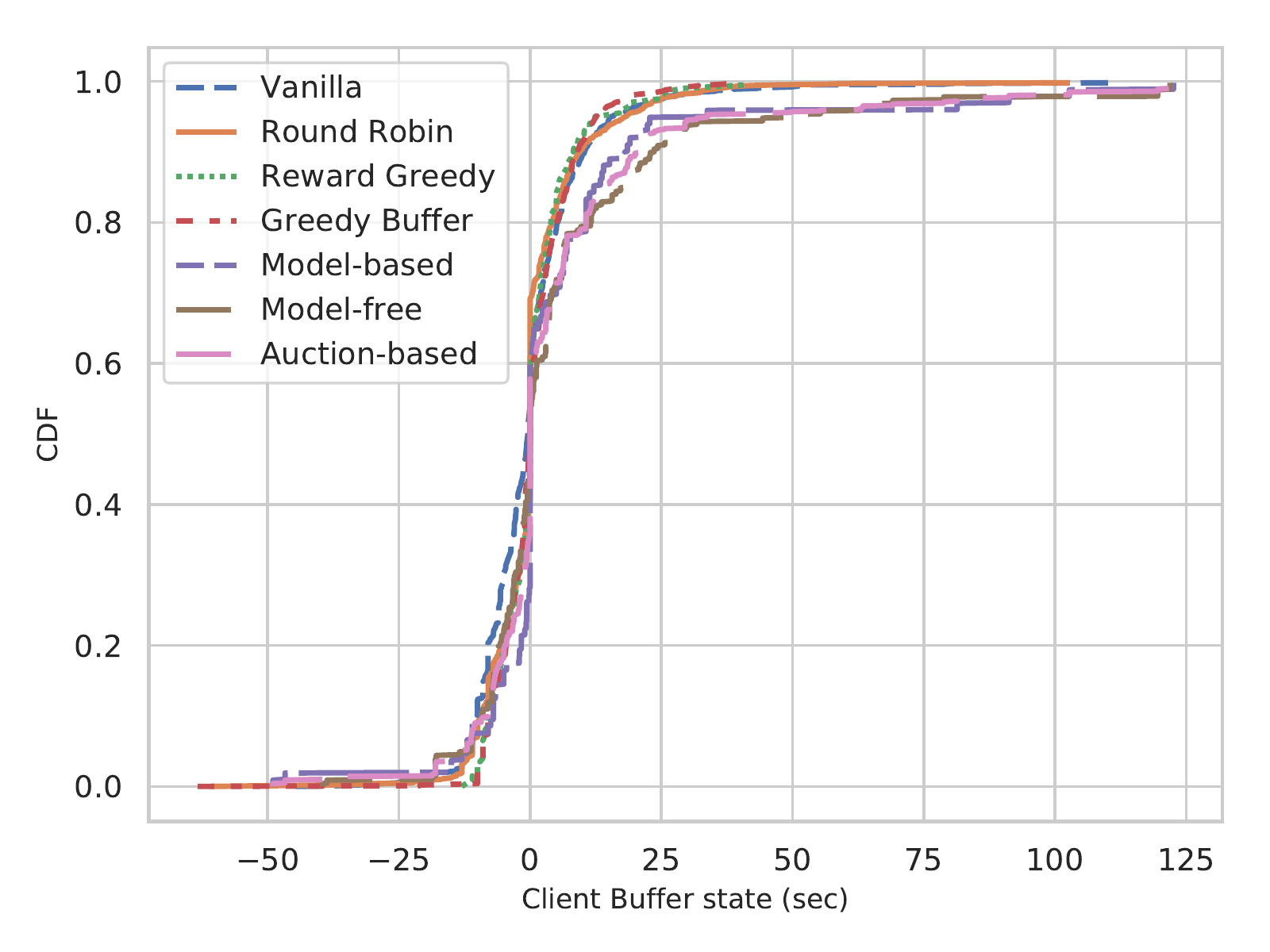}
\caption{Comparison of client Buffer CDF }
\label{fig:buf_cdf}
\end{minipage}\hfill
\begin{minipage}{.32\textwidth}
\centering
\includegraphics[width=1\columnwidth]{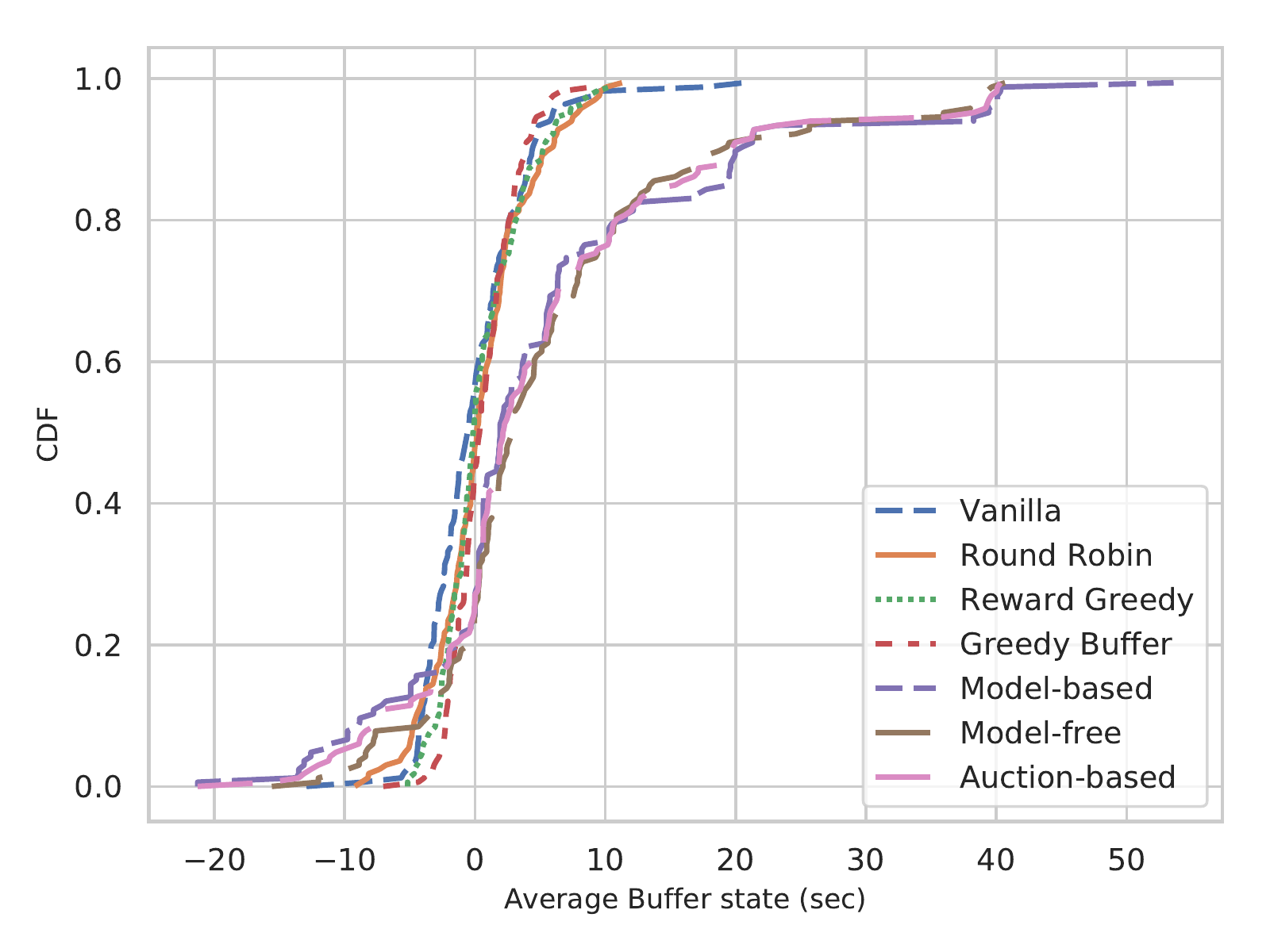}
\caption{Comparison of average Buffer CDF }
\label{fig:buf_avg_cdf}
\end{minipage}
\vspace{-0.1in}
\end{figure*}

An off-the-shelf WiFi router installed with QFlow is used as the Access Point and three Intel NUCs are used to instantiate up to 9 clients (YouTube sessions) for our experiments. Note that each such session can be associated with multiple TCP flows, and we treat all the flows associated with a particular YouTube session identically.  The three NUCs are equipped with 5th generation i7 processors with 8 GB of memory, each capable of running multiple traffic intensive sessions simultaneously. Relevant session information such as ports used by an application, play/load progress, bitrate and stall information for YouTube sessions is collected every second and written to the database.

We use the platform to study several scenarios involving one or more ``bins", each containing two downlink queues, one with a higher bandwidth allocation (resulting in better QoS) than the other using token bucket queueing via Ubuntu Traffic Controller (TC).   As we will see later in Section~\ref{sec:time-varying}, we ensure that devices with similar signal strengths are made to occupy the same bin, and hence do not adversely affect the performance of clients with better signal strengths.   An example with two bins corresponding to "High" and "Low" signal strengths is shown in Figure~\ref{fig:setup}.  A default queue is used for any background traffic.  Two clients may be allowed into each high priority queue.  
For the no differentiation case, we set up a single queue with the same total throughput limit as the sum of all queues in the previous scenarios.  Our control problem is to determine which sessions to assign to which queues.   

\subsection{Policies}
In addition to described model-based, model-free and auction-based policies, we consider four additional policies for choosing these assignments.

{\bf{Vanilla}:} This is the base case with a single queue that is
allocated the full bandwidth, and with no differentiation between clients.

{\bf{Round Robin}:} As the name suggests, we assign clients to the high priority queue in turn.  Although it is computationally inexpensive, work-conserving and prevents starvation, it might lead to the wrong clients (those who have no hope of significantly increasing their QoE) being considered for the high-quality service instead of clients who might benefit much more from the service.

{\bf{Reward Greedy}:}  This policy computes the expected one-step reward on a per-client basis, and assigns clients so as to maximize the sum of rewards.  We can think of this as a myopic version of model-based RL.    This might starve sessions that were unlucky and stalled at some point, since QoE growth rates reduce after stalls.

{\bf{Greedy Buffer}:}
The smooth playout of a video depends on the size of the playback buffer.  When the buffer is empty, the client experiences a stall and the perceived QoE drops. This approach promotes the clients with the lowest buffered video to the high priority queue to prevent this from occurring. This policy might promote the wrong agents who have low buffers because they are at the end of their videos, or those that have stalled multiple times and can never recover high QoE.

\subsection{Static Network Configuration}

In our static configuration, we have just one bin, and each NUC hosts two YouTube sessions to simulate a total of 6 clients.  
The QoE performance comparison of the different policies is shown in Figures \ref{fig:qoe_comp}, \ref{fig:qoe_cdf} and \ref{fig:qoe_avg_cdf}.  We first compare the average QoE of the various policies in the first figure. It is clear that the model-based, model-free and auction-based policies outperform the other policies. This gap in performance becomes even more evident when we compare the CDFs of the individual and the average QoE of the different policies in Figures \ref{fig:qoe_cdf} and \ref{fig:qoe_avg_cdf}. For example, we can observe from Figure \ref{fig:qoe_cdf} that the Model-based, Model-free and Auction-based policies are able to provide a QoE of 5 for almost 90, 85 and 87\% of the time for all clients, whereas it is only about 65\% of the time for the next best policy. Similarly, it can be deduced from Figure \ref{fig:qoe_avg_cdf} that the Model-based, Model-free and Auction-based policies are able to achieve an average QoE of 4.5 for all participating clients in the system about 70, 85 and 90\% respectively. The value of this metric for the next best policy is about 35\%. 

Interestingly, the auction outperforms the model-based policy.  We believe that this is due to the coarse quantization of the state space.  System-wide identification of value is worse affected by such coarse quantization due to the fact that 6 clients together are considered in the sate, whereas in Auction only 1 client is part of the (marginal) state.   Hence, we believe that relative value identification (indexing of client states; see Section~\ref{section:index-policy}) is more accurate in the Auction case.

\begin{figure*}[htbp]
\centering
\begin{minipage}{.32\textwidth}
\centering
\includegraphics[width=1\columnwidth]{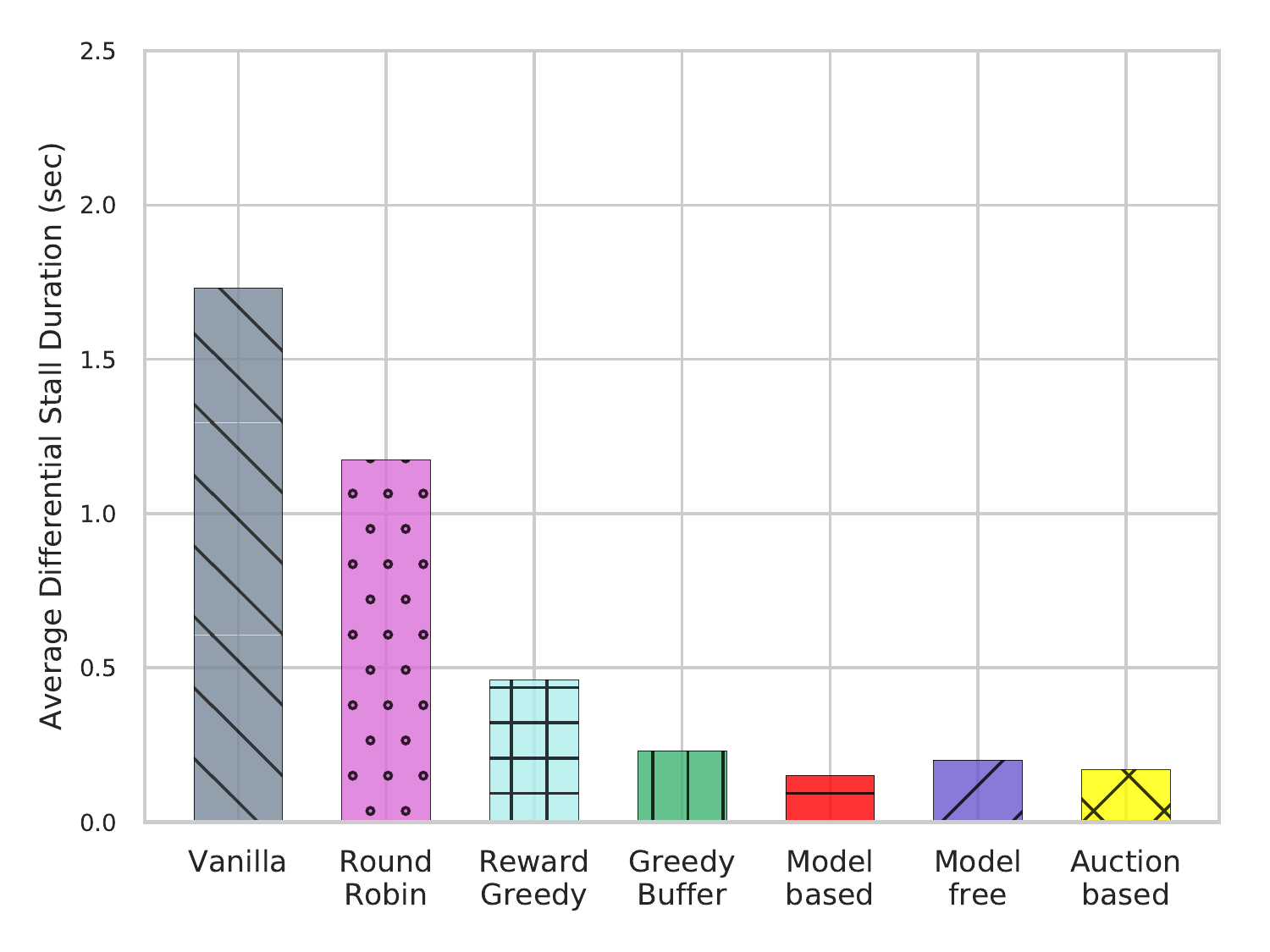}
\caption{Comparison of average stall duration }
\label{fig:stall_comp}
\end{minipage}\hfill
\begin{minipage}{.32\textwidth}
\centering
\includegraphics[width=1\columnwidth]{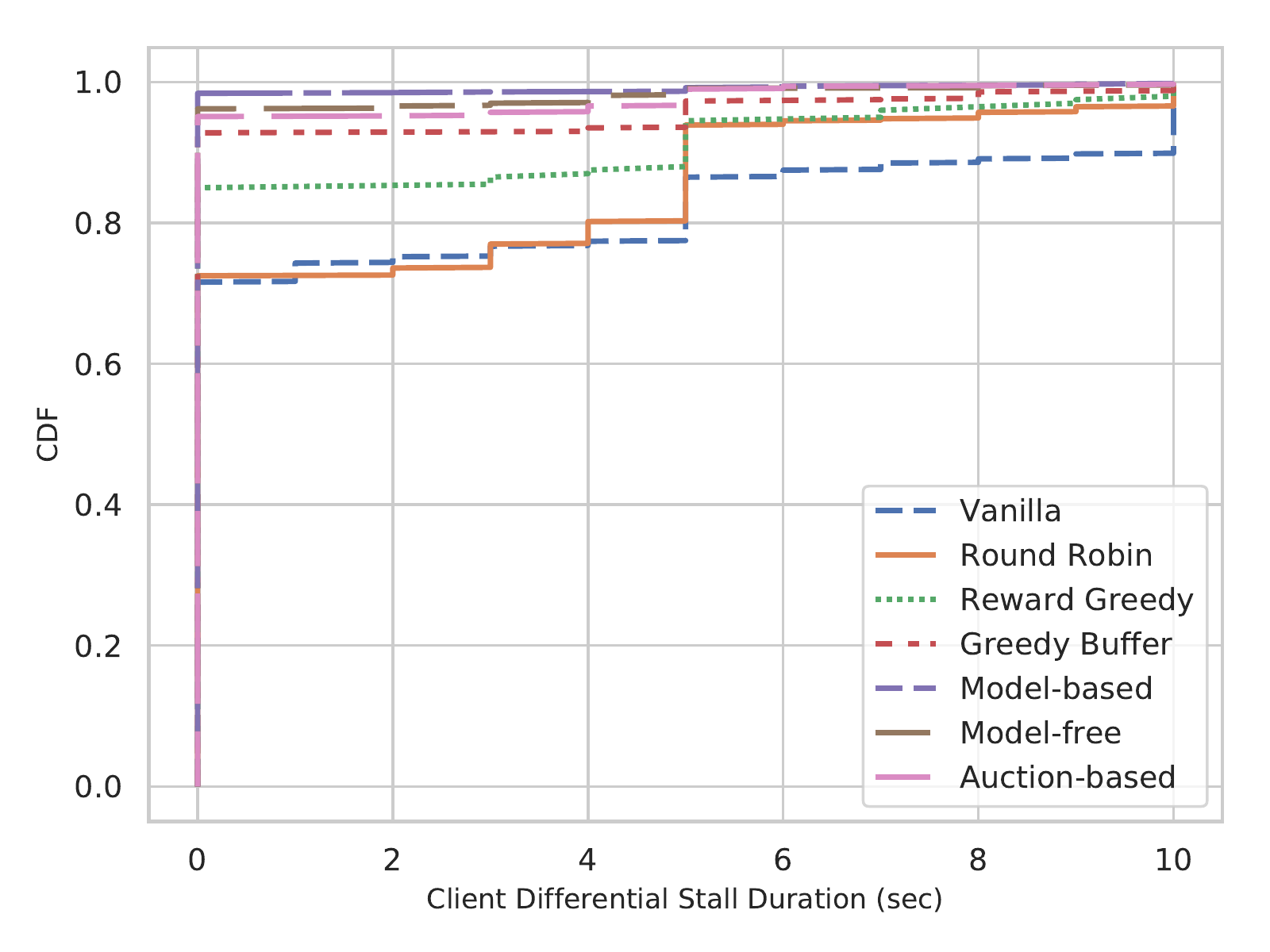}
\caption{Comparison of stall duration CDF }
\label{fig:stall_cdf}
\end{minipage}\hfill
\begin{minipage}{.32\textwidth}
\centering
\includegraphics[width=1\columnwidth]{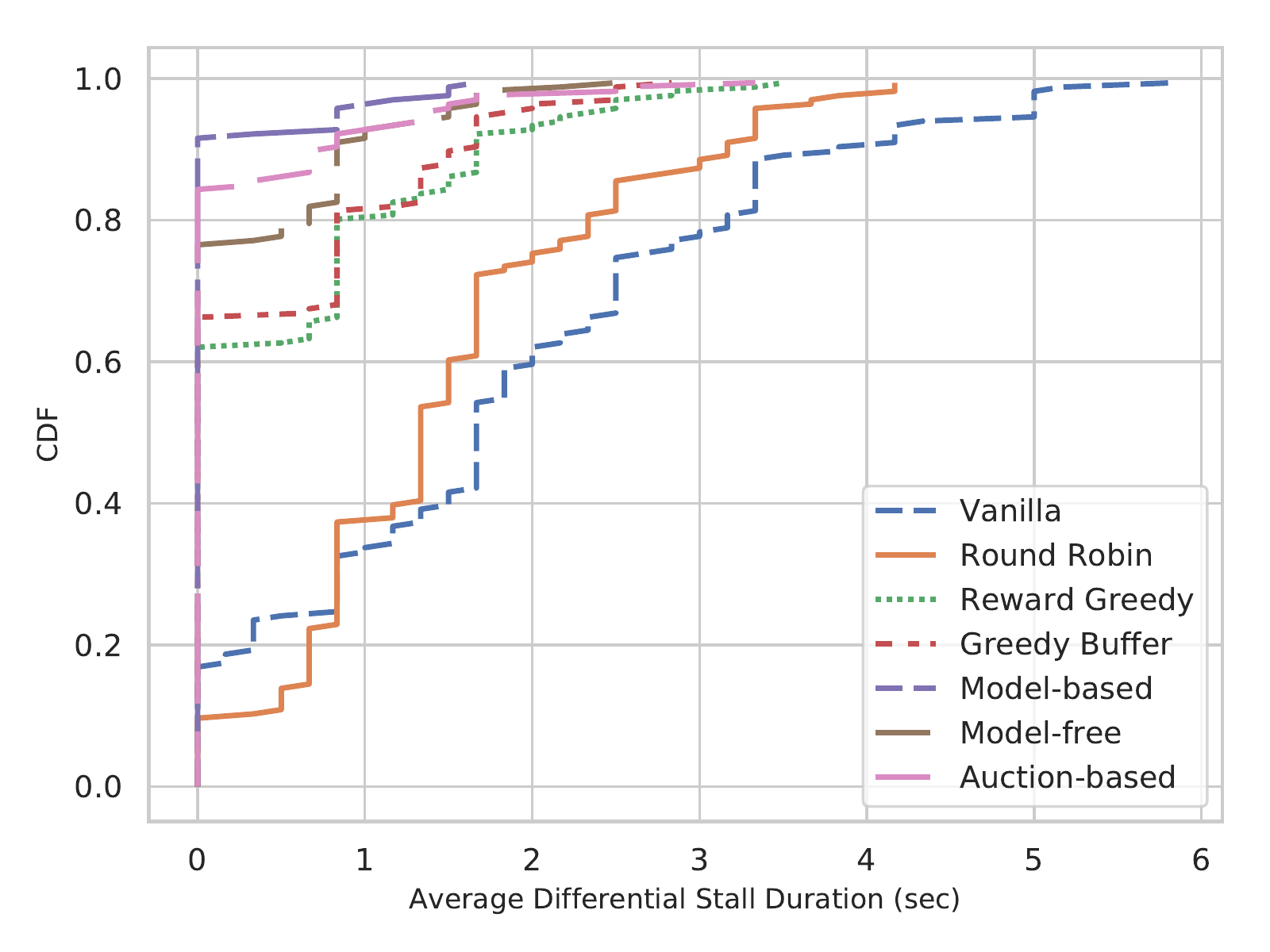}
\caption{Comparison of average stall duration CDF }
\label{fig:stall_avg_cdf}
\end{minipage}
\begin{minipage}{.32\textwidth}
\centering
\includegraphics[width=1\columnwidth]{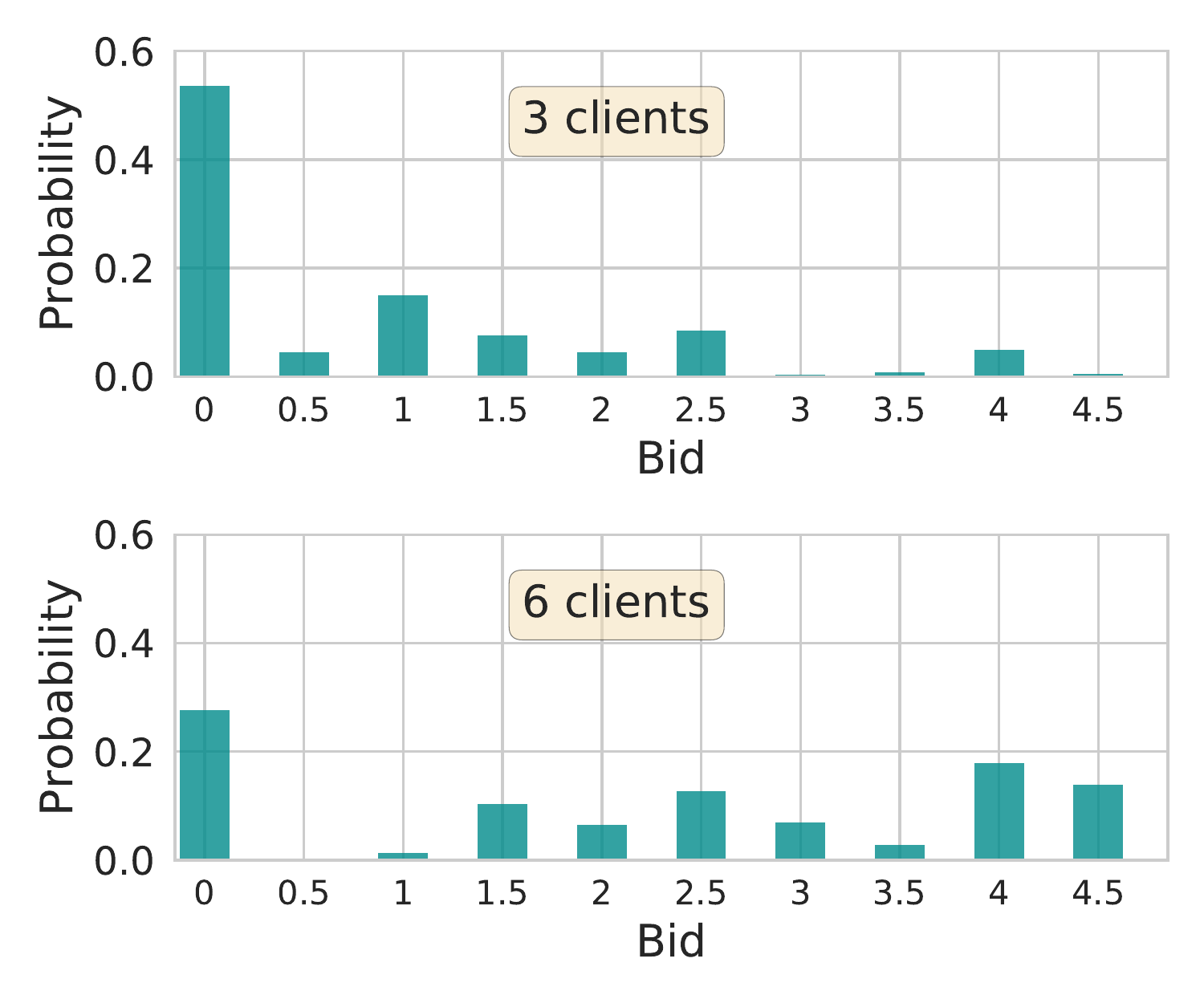}
\caption{Bid distribution for 6 and 3 client configurations}
\label{fig:bid_dist}
\end{minipage}\hfill
\begin{minipage}{.32\textwidth}
\centering
\includegraphics[width=1\columnwidth]{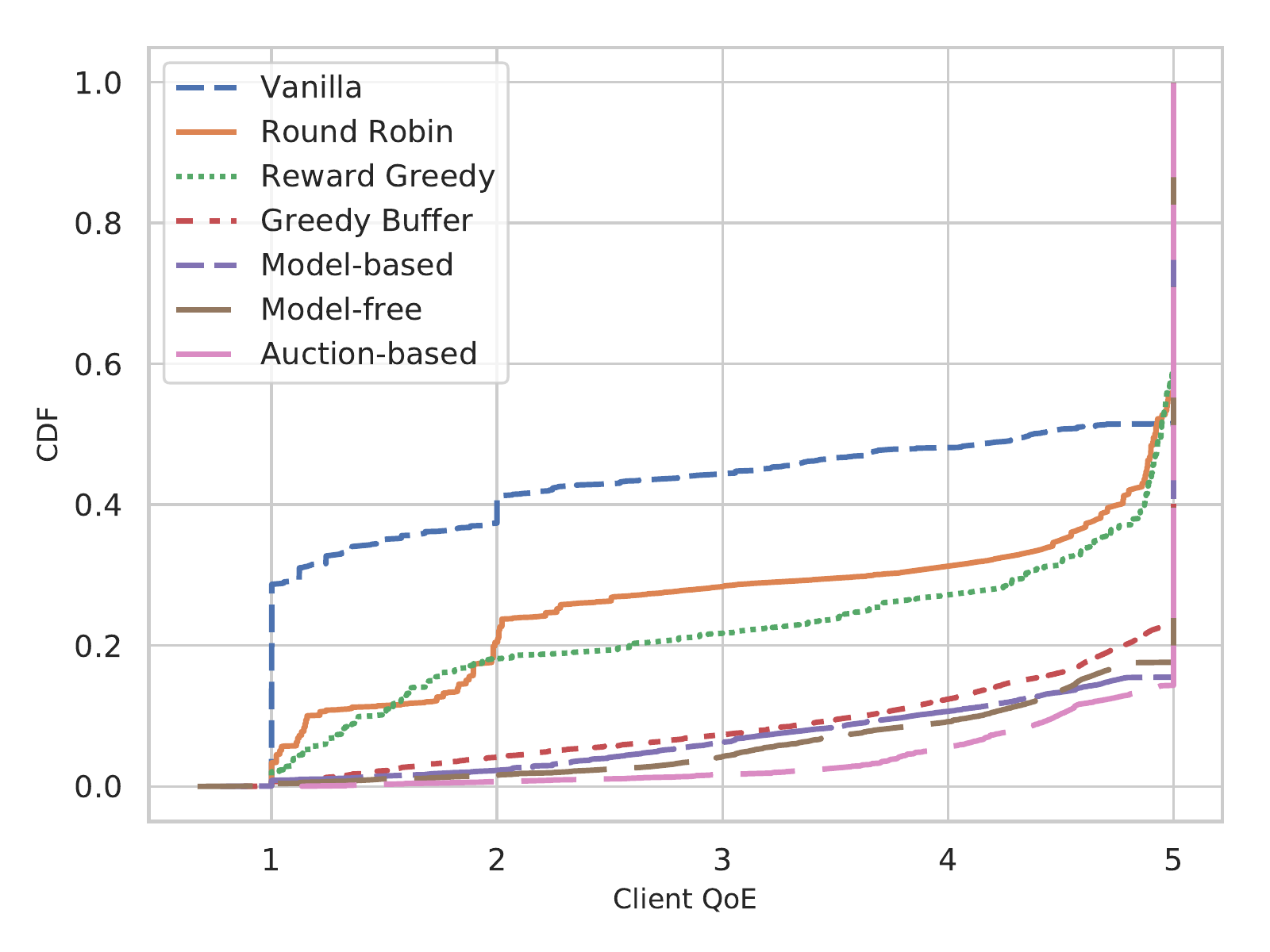}
\caption{Comparison of client QoE CDF for dynamic clients}
\label{fig:qoe_m_cdf}
\end{minipage}\hfill
\begin{minipage}{.32\textwidth}
\centering
\includegraphics[width=1\columnwidth]{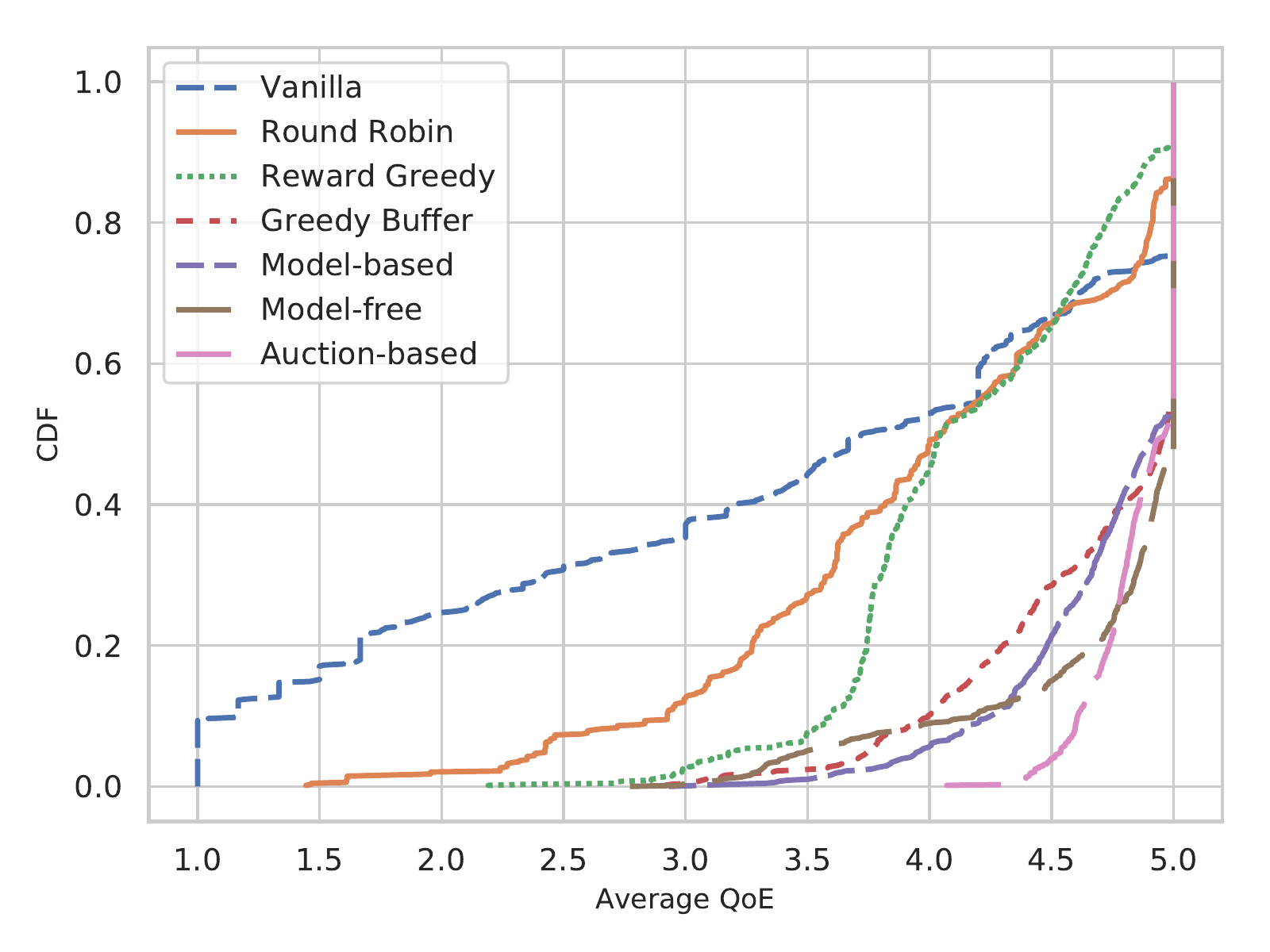}
\caption{Comparison of average QoE CDF for dynamic clients}
\label{fig:qoe_m_avg_cdf}
\end{minipage}
\begin{minipage}{.32\textwidth}
\centering
\includegraphics[width=1\columnwidth]{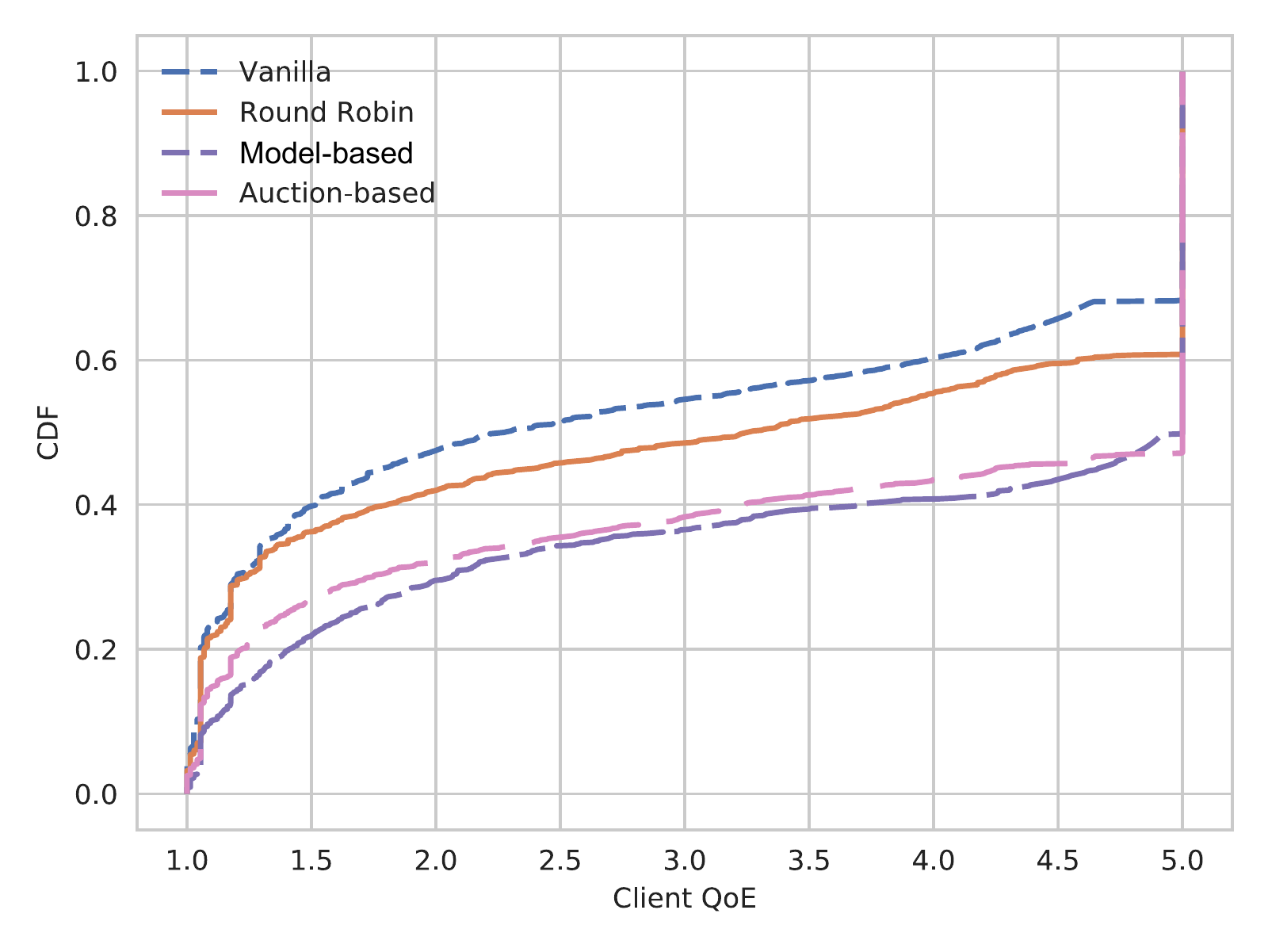}
\caption{Comparison of QoE CDF for Bad Channel}
\label{fig:qoe_cdf_bad}
\end{minipage}\hfill
\begin{minipage}{.32\textwidth}
\centering
\includegraphics[width=1\columnwidth]{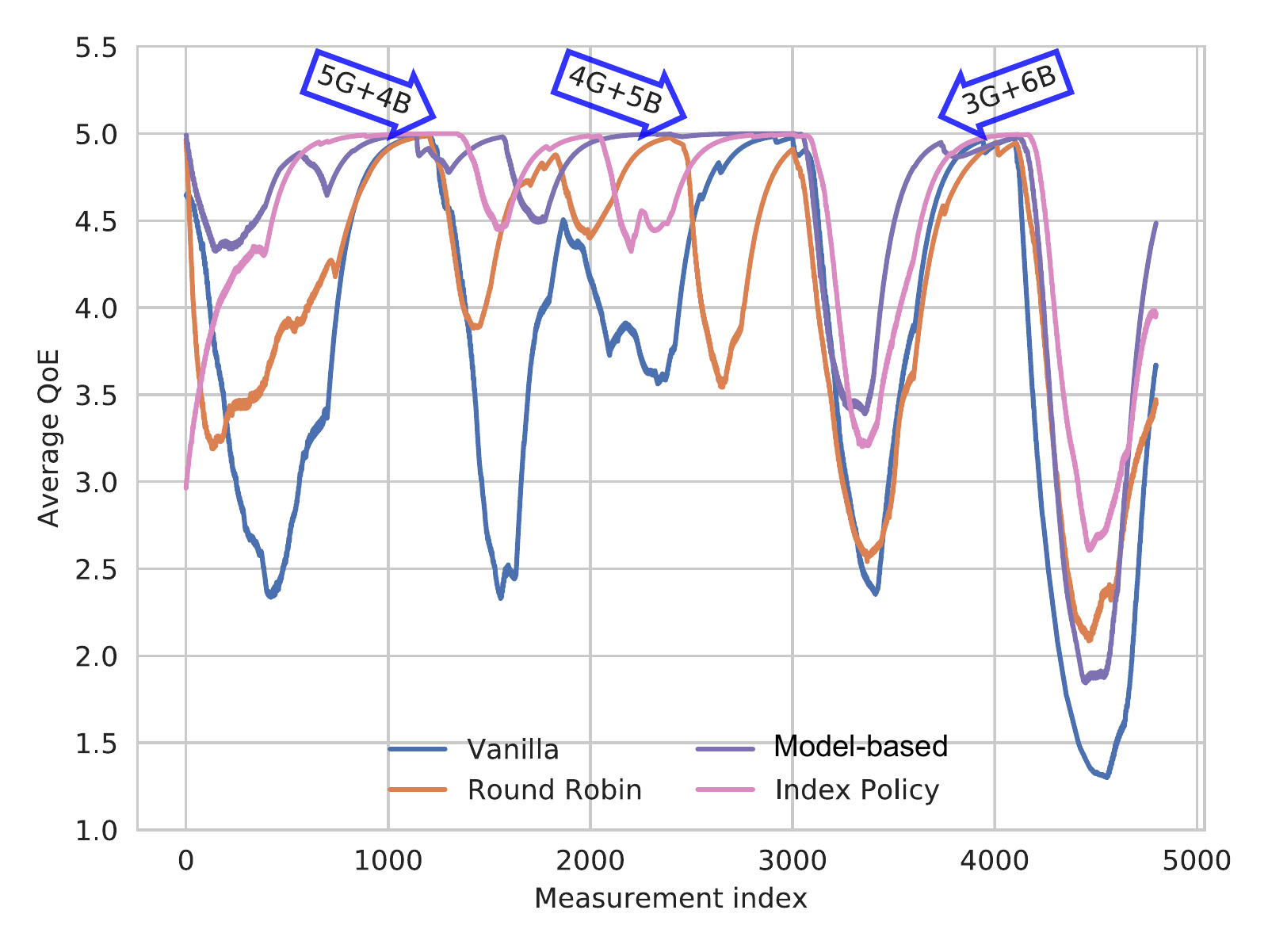}
\caption{Evolution of QoE: Dynamic clients with variable channels}
\label{fig:qoe_evol}
\end{minipage}\hfill
\begin{minipage}{.32\textwidth}
\centering
\includegraphics[width=1\columnwidth]{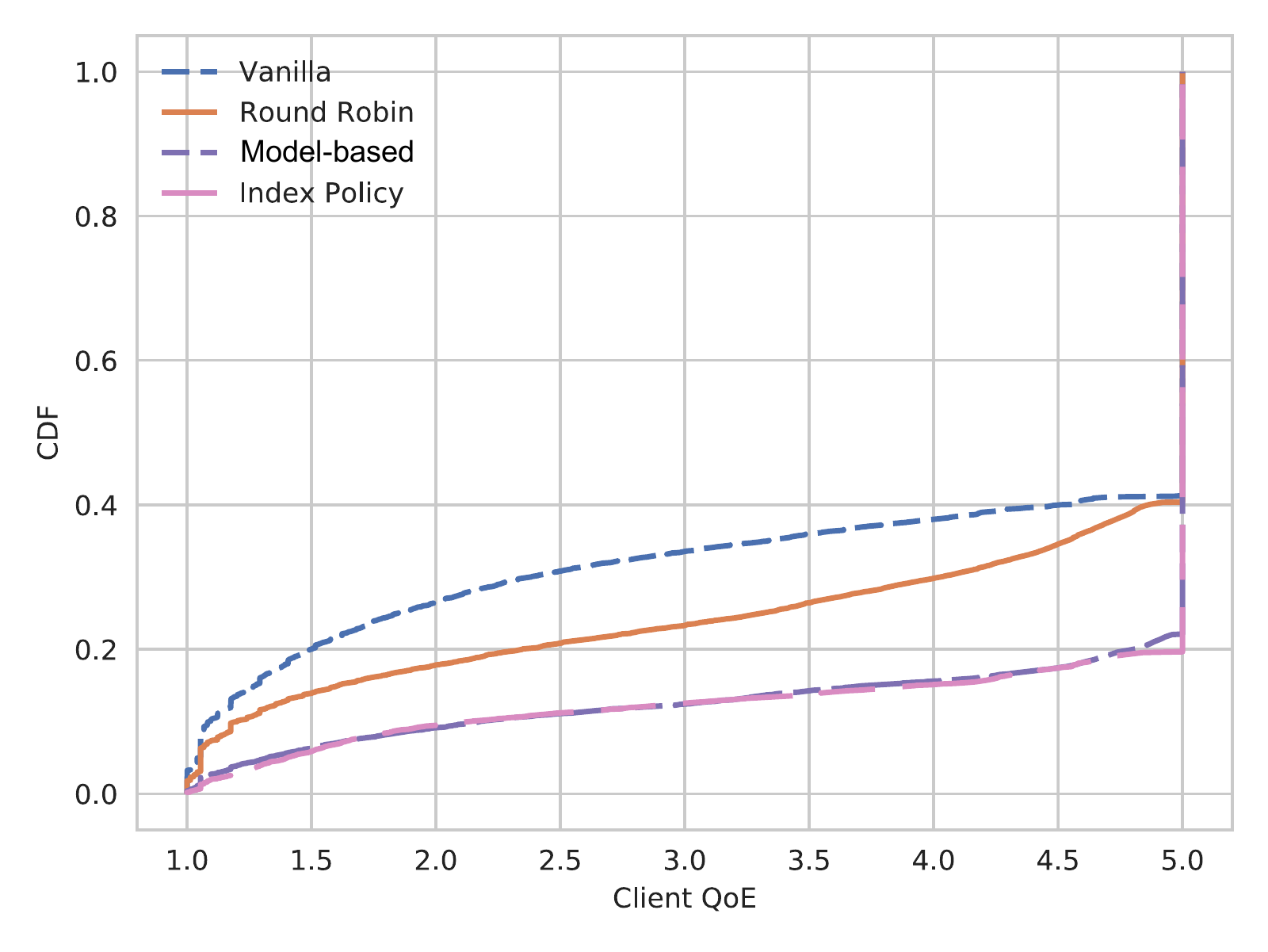}
\caption{Comparison of QoE CDF for dynamic clients with variable channels}
\label{fig:qoe_channel_cdf}
\end{minipage}\hfill
\vspace{-0.1in}
\end{figure*}

The QoE experienced by a client is affected by the buffer state of the client and the stalls experienced during video playback. Hence, we study the buffer state and the stall durations experienced by the clients under the different policies. Similar to the QoE plots, we compared the averages, the CDFs of the individual and the average values for both these features in Figures \ref{fig:buf_comp} to \ref{fig:stall_avg_cdf}. Again, it is evident from the figures that the Model-based, Model-free and Auction-based policies ensure better buffer state and lower stall durations (both individual and average) than the other policies under consideration.

We also compared the bid distributions of the clients in the Auction-based policy for two different client configurations. The first configuration had 6 clients whereas the second one had 3, with the total bandwidth allocation kept the same. The comparison of the two distributions is shown in Figure~\ref{fig:bid_dist}.  When there are more clients participating, resources are scarce and valuable, and so, clients tend to bid higher in order to get into the high priority queue and experience better QoE.  When the total number of clients is low, everyone experiences good QoE irrespective of the queue they are assigned to, and there is no incentive to bid higher.

\subsection{Dynamic Number of  Clients}
We next study the performance of the policies in a scenario with a varying number of clients.  We still maintain a single bin, but choose the number of active clients in the system to vary between 4 and 6, while keeping the bandwidth allocation same as that of the static configuration.  We study there policies, namely (i) Model-free: This is obtained by retraining Q-Learning for 4, 5 and 6 clients, (ii) Model-based: This is obtained by retraining the model-based approach for 4, 5 and 6 clients, (iii) Auction-based (index policy): This is obtained by training via the auction for 6 clients, and using the ordering of values so obtained as state indices for any other number of clients.

We consider a larger timescale of 30 minutes for changing the number of clients participating in the system.  We start with 6 clients in the system and then remove 1 client each for the next two time periods. At the end of the third period, we introduce two more clients in the system. It is observed that Model-based, Model-free and Auction-based (index) policies perform well irrespective of the number of users in the system, whereas other policies only do well when there are relatively fewer clients in the system.  As before, we note that the Auction-based (index) policy performs slightly better than the other two RL policies.  Also, as we pointed out earlier, the index approach also implies that we only need train once, and not for each possible number of clients separately.  Hence, it has a lower complexity as compared to the other two approaches.

Observe that since the bandwidth allocation is the same, reducing the number of clients implies relaxation of the resource constraints and hence, other policies see an improvement in performance. This can be seen in Figures \ref{fig:qoe_m_cdf} and \ref{fig:qoe_m_avg_cdf}, where the CDF curves of the other policies are closer to those of model-based, model-free and auction-based policies. Even so, these three policies exhibit the best performance, which reinforces their superiority over other policies in both static and dynamic client scenarios.

\subsection{Time Varying Channel Conditions}
\label{sec:time-varying}

Wireless clients could have time varying signal strengths, and consequently face different link-level throughputs, latencies, and loss rates.  We need to ensure that clients having lower signal strengths do not adversely affect the performance of clients with better signal strengths by occupying the channel longer for each packet transmission~\cite{heusse2003performance}.  Hence, we create two bins of downlink queues, each containing a high priority and a low priority queue as shown in Figure~\ref{fig:setup}.   We then have a $Good$ bin for clients with high signal strengths, and a $Bad$ bin for those who have low signal strengths. 

In order to ensure repeatably of experiments across different polices as clients experience good and bad channels, we emulate a bad channel by reducing the throughput, and increasing the latency and loss rates of the queues in the $Bad$ bin as compared to those in the $Good$ bin using Ubuntu Network Emulator (NetEm). We then create a sequence of good and bad (emulated) channel conditions over time for each client that we repeat for each policy.  In order to determine a realistic emulation of what ``bad'' might mean for video streaming, we ran several hours of experiments with clients having low signal strengths (via antenna attenuators) to determine appropriate emulator settings. Thus, we are able to mimic varying network conditions by dynamically assigning the sessions hosted on the NUCs to either the $Good$ or the $Bad$ bin.

In what follows, we consider four policies: two baselines, namely (i) Vanilla and (ii) Round Robin, and two advanced polices that show good performance, namely (iii) Model-based RL and (iv) Auction-based (which yields an ordering for the index policy).   We first illustrate the difference in achieved performance for the different policies with a static 6 clients under $Good$ vs. $Bad$ channel conditions by comparing Figures \ref{fig:qoe_cdf} and \ref{fig:qoe_cdf_bad}. 
We observe that the gap between the baseline and advanced policies decreases in the $Bad$ channel scenario, but Model-based RL and the Auction-based policies still achieve higher QoE for the clients.  

We next fix a sequence of client configurations (number of active clients) under each channel condition for the evaluation of all policies. The first configuration consists of 6 clients under $Good$ channel conditions and 3 under $Bad$ channel conditions. We decrease the number of clients in the $Good$ channel by 1 and increase those in the $Bad$ channel by 1 for the next three intervals. The evolution of the average QoE for each of the policies for the above sequence is shown in Figure \ref{fig:qoe_evol}. The Model-based RL and Auction-based (index) policies exhibit a high average QoE in most of the configurations except for the last where it is not possible to achieve a high QoE for the 6 clients in the $Bad$ channel. Even in such a scenario, the drop in QoE is less severe than the other policies. 

Finally, we show the overall CDF of client QoEs taken over the whole experiment interval in Figure~\ref{fig:qoe_channel_cdf}.  While the QoE improvement from the baseline using the learning-based policies is not as striking as it is in Figure~\ref{fig:qoe_cdf}, the QoE samples with the learning policies have a perfect QoE score in about 80\% of the samples as compared to he baseline policies that only manage this in about 60\% of the samples.

\section{Conclusion}
\label{sec:conclusion}
In this paper, we considered the design, development and evaluation of QFlow, a platform for learning based edge network configuration. 
Working with off-the-shelf hardware and open source operating systems and protocols, we showed how to couple queueing, learning, markets and scheduling to develop a system that is able to reconfigure itself to best suit the needs of video streaming applications. As our YouTube observations suggest, such a holistic framework that accounts for this entire chain can reveal efficiencies and interactions that a narrow focus on individual components of the system is incapable of achieving. 

We instantiated  a variety of policies on the platform.  We began with model-free and model-based RL, and showed that the model needed in the latter case is the marginal transition kernel of the system, which is learned quite easily.  We also showed that using an auction framework is able to a elicit truthful proxy for state in terms of the bid made for prioritized service.  We also discovered ordering of state values that can be applied directly as a simple index policy.  The existence of such a policy suggests that perhaps we can directly design an RL approach to find the index policy, and this will be considered in the future.    
%
We believe that the application of our system will be in upcoming small cell wireless architectures such as 5G, and our goal will be to extend our ideas to such settings.

\bibliographystyle{IEEEtran}
\bibliography{references}

\end{document}